\documentclass[10pt,journal]{IEEEtran}
\usepackage{color}
\usepackage{xcolor}
\usepackage{cite}
\usepackage{setspace} 
\usepackage{amsfonts} 
\usepackage{amsmath,amsthm,amssymb}
\usepackage{multirow}
\usepackage{hhline}
\usepackage{mathtools}
\usepackage{amsmath}
\usepackage{multirow}
\usepackage[normalem]{ulem}
\usepackage{soul}
\usepackage{verbatim}
\usepackage{combelow}

\usepackage[T1]{fontenc}
\usepackage{graphicx}
\usepackage{textcomp}
\def\BibTeX{{\rm B\kern-.05em{\sc i\kern-.025em b}\kern-.08em
    T\kern-.1667em\lower.7ex\hbox{E}\kern-.125emX}}
\usepackage[font=small]{caption}
\usepackage{breqn}
\usepackage{array}
\newcolumntype{P}[1]{>{\centering\arraybackslash}p{#1}}
\newcolumntype{M}[1]{>{\centering\arraybackslash}m{#1}}

\usepackage{colortbl}
\usepackage{algorithm}
\usepackage{algorithmicx}
\usepackage{algpseudocode}
\usepackage{lipsum}

\definecolor{inchworm}{rgb}{0.7, 0.93, 0.36}

\usepackage{booktabs}
\usepackage[T1]{fontenc} 
\usepackage[utf8]{inputenc}
\usepackage{hyperref}

\begin{document}
\title{SCAR: State-Space Compression for AI-Based Radio Resource Management of Vehicular Services}

\author{Ioan-Sorin Com\textcommabelow{s}a,
        Purav Shah,
        Karthik Vaidhyanathan,
        Deepak Gangadharan,
        Christof Imhof, \\
        Per Bergamin,
        Aryan Kaushik,
        Gabriel-Miro Muntean,
        and
        Ramona Trestian 
\vspace{-0.05in}

\IEEEcompsocitemizethanks{\IEEEcompsocthanksitem 
I.-S. Com\textcommabelow{s}a, C. Imhof, and P. Bergamin are with the Institute for Research in Open-, Distance- and eLearning, Swiss Distance University of Applied Sciences, Brig, CH-3900, Switzerland (e-mails: \{ioan-sorin.comsa, christof.imhof, per.bergamin\}@ffhs.ch). 
P. Shah and R. Trestian are with London Digital Twin Research Centre, Middlesex University London, NW4 4BT, Hendon, London, U.K. (e-mails: \{r.trestian, p.shah\}@mdx.ac.uk). 
K. Vaidhyanathan is with the Software Engineering Research Center, IIIT-Hyderabad, Hyderabad 500032, India. (e-mail: karthik.vaidhyanathan@iiit.ac.in). 
D. Gangadharan is with the Computer Systems Group, IIIT Hyderabad, Hyderabad 500032, India. (e-mail: deepak.g@iiit.ac.in). 
A. Kaushik is with OUF Innovative Ltd., London, U.K., e-mail: (a.kaushik@ieee.org).
G.-M. Muntean is with School of Electronic Engineering, Dublin City University, D09 V209, Dublin, Ireland (e-mail: gabriel.muntean@dcu.ie).
}
}

\vspace{-0.05in}
\IEEEtitleabstractindextext{%
\begin{abstract}
The increasing demand for connected vehicular services poses significant challenges for AI-based network and service management due to the high volume and rapid variability of network state information. Traditional management and control mechanisms struggle to scale when processing fine-grained metrics such as Channel Quality Indicators (CQIs) in dynamic vehicular environments. To address this challenge, we propose SCAR (State-Space Compression for AI-Based Radio Resource Management), an edge-assisted framework that abstracts high-dimensional CQI information into compact representations for AI-based management. SCAR employs machine-learning (ML)–based compression techniques, including clustering and radial basis function (RBF) networks, to reduce the dimensionality of CQI-derived state information while preserving essential features relevant to management decisions. The resulting compressed states are used to train reinforcement learning (RL)–based management policies that dynamically adapt the throughput–fairness trade-off according to service-level fairness objectives defined by the NGMN. Simulation results show that CQI compression improves average NGMN feasible-region occupancy from 56.32\% to 88.07\% across ten RL algorithms, reaching 97.76\% for the best-performing RL configuration, with limited throughput degradation and competitive computational overhead. Furthermore, simulated annealing with stochastic tunneling (SAST)-based clustering reduces state compression distortion by approximately 10\% compared with conventional K-means clustering. These results demonstrate SCAR's ability to provide compact state representations for RL-based network and service management while improving service-level fairness in dynamic vehicular systems.
\end{abstract}

\vspace{-0.05in}
\begin{IEEEkeywords}
Network and Service Management, Radio Resource Management,  Channel Quality Indicator, State-Space Compression, Reinforcement Learning, Vehicular Networks.
\vspace{-0.05in}
\end{IEEEkeywords}}
\maketitle
\IEEEdisplaynontitleabstractindextext
\IEEEpeerreviewmaketitle

\vspace{-0.1in}
\section{Introduction}\label{sec:introduction}
The evolution of intelligent, software-driven mobile networks has increased the demand for adaptive network and service management \cite{sharma2023toward}. Beyond mission-critical Ultra-Reliable Low-Latency Communication (URLLC) services \cite{8636206}, high-throughput and delay-sensitive vehicular applications, including infotainment, cloud gaming, augmented reality (AR), and video streaming, require scalable management under dynamic network conditions \cite{10818589,yeznabad2024qoe}. Connected vehicles generate extensive network measurements, including Channel Quality Indicators (CQI), Signal-to-Noise Ratio (SNR), and latency information, whose efficient processing is essential for maintaining service quality and fairness.

However, high-dimensional CQI reports increase computational and control-plane overhead, potentially limiting real-time responsiveness \cite{9026965}. Mobility further complicates computation offloading and resource management in Vehicular Edge Computing (VEC) \cite{10049431}, 
motivating adaptive edge-based optimization \cite{10330753} and workload-aware resource management \cite{10571962}.
Conventional resource management based on centralized heuristics struggles with rapid network-state variations \cite{8926512}. Machine learning (ML), particularly reinforcement learning (RL), enables adaptive routing, resource allocation, and policy optimization \cite{casas2020intelligent,zhang2022reinforcement,comsa_thesis_2014}, including throughput--fairness management in vehicular networks \cite{yuan2021harnessing}. Nevertheless, high-dimensional measurement inputs remain a scalability challenge for RL-based management, motivating compression-aware state representations.

To address these challenges, we propose SCAR (State-Space Compression for AI-Based Radio Resource Management), an edge-assisted framework integrating CQI state-space compression with RL-based control. SCAR employs K-means clustering to identify representative CQI patterns and online Radial Basis Function Network (RBFN) classification to map incoming reports into compact states. An RL-based controller then dynamically adapts the throughput--fairness trade-off according to Next Generation Mobile Networks (NGMN) service-level fairness objectives. By reducing the dimensionality of CQI observations, SCAR enables RL-based scheduling with compact state
representations while improving compliance with NGMN
service-level fairness constraints across different compression configurations and RL algorithms.

\vspace{-0.2in}
\subsection{Related Work}\label{sec:related_works}
Efficient classification and abstraction of network-state information are essential for scalable ML-driven network management, particularly in dynamic vehicular environments. K-means clustering is widely employed due to its simplicity and effectiveness in minimizing intra-cluster distances \cite{unsupervised_learning_2019}. Its applications include mmWave multipath partitioning \cite{8734258}, cooperative spectrum sensing \cite{6635250}, user behavior analysis \cite{7811244}, and content caching \cite{8114341}. However, K-means is sensitive to initialization and local optima, motivating improved seeding and optimization strategies \cite{mount_sa}. Moreover, these applications do not explicitly address CQI state-space abstraction for RL-based vehicular network management.

Supervised learning techniques, including k-nearest neighbors, decision trees, Bayesian classifiers, and support vector machines, have been applied to radio resource management (RRM) \cite{7289481}. Neural networks have also supported cellular traffic prediction \cite{8667446} and 
enhancing network capacity under varying network loads
\cite{8337851}. 
Hybrid approaches combining supervised and unsupervised learning techniques, as well as game-theoretic optimization models, have been investigated for content dissemination and interference management \cite{8574942,8474384}. However, computational requirements during online inference can constrain their applicability to time-sensitive network management. Radial Basis Function Networks (RBFNs), which support efficient classification and are compatible with clustering-based representations \cite{neural_networks_1999}, provide a suitable mechanism for mapping high-dimensional CQI measurements into compact network-state representations.

Reinforcement learning (RL) has been extensively investigated for adaptive network and service management, including distributed resource allocation and coordination across network domains \cite{zhang2022distributed}. Learning-based approaches have also been proposed for microservice placement and migration in multi-access edge computing \cite{ray2023learning}, while attention-based deep RL has been investigated for resource management under complex network conditions \cite{chang2023attention}. These studies demonstrate the benefits of adaptive control but generally focus on policy optimization using available state information rather than reducing the dimensionality of measurement-driven RL inputs.

Recent research has further advanced intelligent resource management in heterogeneous edge environments. Multi-objective optimization integrating multi-agent learning, UAV-assisted communications, and intelligent wireless architectures has been explored to improve throughput and resource utilization \cite{https://doi.org/10.1002/ett.70210}. Adaptive deep RL and traffic-aware network slicing have been investigated for dynamic computation offloading \cite{10756668}. Hierarchical federated learning and graph-enhanced multi-agent approaches have also supported distributed offloading and digital-twin migration under mobility \cite{10.1504/ijsnet.2025.148455}, while graph-attentive multi-agent RL has been applied to SLA-aware service orchestration \cite{11430686}. In vehicular edge networks, trajectory-aware multi-objective optimization has considered mobility, communication and computation delays, energy consumption, and load balancing for task offloading \cite{11197650}. Although these approaches improve adaptive resource management, they typically assume that the underlying network-state information is already available in a suitable representation for decision-making.

The increasing demand for high-throughput, delay-sensitive vehicular applications further motivates efficient network-state processing. Multi-access Edge Computing (MEC) enables processing closer to vehicles, supporting reduced latency and improved service quality \cite{9040539}. Cooperative data sharing and edge-assisted perception have also demonstrated the importance of compression and 
federated learning in autonomous systems \cite{9606821}. Nevertheless, these advances primarily address computation, communication, and service execution, while the construction of compact CQI-based state representations for learning-driven network management remains less explored.

Existing CQI compression techniques have primarily targeted signaling overhead rather than the dimensionality of RL state representations. Selective feedback mechanisms transmit CQI according to scheduling-probability thresholds \cite{4257444}, while K-means-based approaches reduce congestion through pattern clustering \cite{8292311}. Deep learning techniques have also been investigated for CQI-related compression in massive MIMO systems \cite{8482358}, and 
average CQI values across the system bandwidth have been employed for wideband scheduling \cite{7410050}. Although these methods reduce feedback requirements or simplify channel-state information, they do not explicitly integrate CQI compression with RL-based fairness management. In contrast, SCAR combines offline K-means clustering, online RBFN classification, and RL-based control to construct compact CQI representations for adaptive throughput--fairness management in dynamic vehicular environments.

\begin{figure*}[t]
\centering
\includegraphics[width=14.5cm]{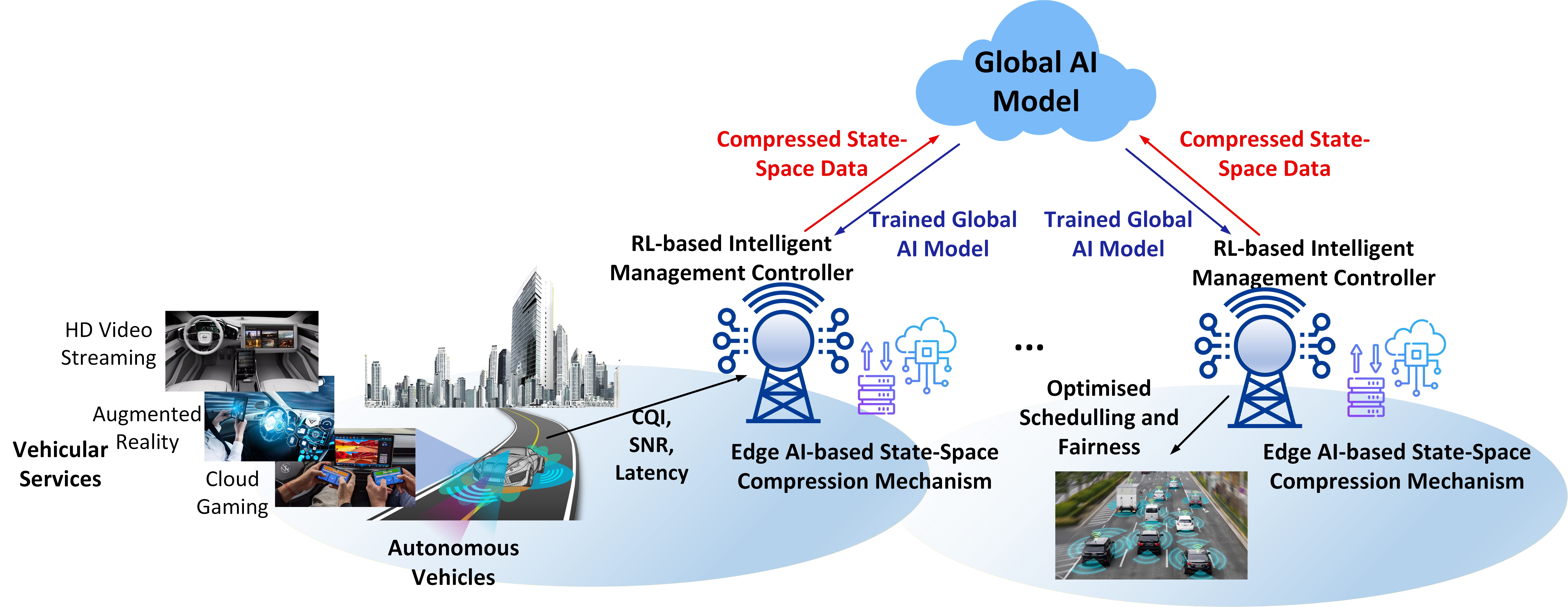}
\caption{Proposed SCAR Framework}
\label{figure_1}
\centering
\vspace{-0.2in}
\end{figure*}

\vspace{-0.1in}
\subsection{Contributions}\label{sec:contributions}
This paper proposes \textbf{SCAR} (\textbf{S}tate-Space \textbf{C}ompression for \textbf{A}I-Based Radio \textbf{R}esource Management), a framework addressing the high dimensionality of network-state information used in RL-based management of dynamic vehicular networks. By abstracting CQI-derived information at the network edge, SCAR provides compact state representations for learning-based control. The main contributions are as follows:

\noindent \textit{1) State-Space Compression via Offline Clustering Optimization}: We introduce a Simulated Annealing with Stochastic Tunneling (SAST)-enhanced K-means clustering method to construct representative state prototypes from CQI datasets. This optimization improves clustering quality by mitigating initialization sensitivity and convergence to local optima, supporting robust state abstraction for downstream learning.

\noindent \textit{2) Computationally Efficient Online State Classification at the Edge}: An online RBFN classifier enables real-time classification of incoming CQI-derived states using precomputed clusters. Its lightweight inference and generalization capabilities support efficient state recognition for edge-assisted network management under dynamic vehicular conditions.

\noindent \textit{3) Compression-Aware Reinforcement Learning for RRM}: Compressed state representations are integrated into an RL-based management controller to guide service allocation under fairness constraints. Results show that compression-aware RL increases the time spent in the NGMN-defined feasible region by over 15\% compared with RL approaches using uncompressed states, while maintaining competitive computational runtime. The trade-off analysis further demonstrates that this improvement is achieved with limited throughput degradation, while balancing quantitative fairness and compliance with the NGMN fairness requirement.

SCAR transforms bandwidth-dependent CQI measurements into fixed-dimensional state representations, supporting potential integration with different resource-management mechanisms and network architectures. Its applicability across various bandwidths and radio access technologies in heterogeneous vehicular communication environments remains subject to future validation.

\vspace{-0.1in}
\subsection{Paper Organization}\label{sec:organisation}
The rest of the paper is organized as follows: Section II presents the proposed SCAR framework and the RL-based management model. In Section III, the proposed CQI compression scheme based on K-means algorithm and RBFN classification is detailed. Section IV analyzes the simulation results for the SCAR framework and shows the advantage of using the proposed compression framework in RL-based management systems. Finally, Section V concludes the paper.

\vspace{-0.15in}
\section{Proposed SCAR Framework}\label{section: scar_framework}
\vspace{-0.05in}
The SCAR framework, illustrated in Fig. \ref{figure_1}, supports adaptive AI-assisted network and service management in dynamic vehicular environments through state-space compression. Its core components comprise (1) edge-assisted CQI compression and (2) RL-based management policy control. By processing high-dimensional CQI measurements at the network edge, SCAR constructs compact state representations for learning-based management under varying user populations and channel conditions. The architecture also illustrates a potential extension toward distributed model refinement across edge and cloud domains, which remains outside the scope of the present experimental evaluation.

SCAR employs a two-stage machine learning pipeline for CQI compression. First, K-means clustering enhanced with the SAST algorithm identifies representative CQI patterns. Subsequently, an RBFN classifier maps incoming CQI reports into compact, fixed-dimensional state representations. These representations are provided to an RL-based management controller that dynamically adjusts resource allocation parameters to regulate the throughput-fairness trade-off according to NGMN-defined service-level objectives. Clustering optimization, RBFN training, and RL policy training are performed offline, while online operations involve CQI classification and RL policy inference for fairness-aware RRM decisions, separating computationally intensive model preparation from runtime control.

As a potential extension, SCAR could support distributed model refinement through periodic exchanges of compressed CQI representations and learned parameters between edge nodes and a centralized coordination entity. Such synchronization could operate at a slower timescale than the online management loop, allowing clustering, classification, and RL models to be periodically updated without requiring centralized real-time control. The implementation and evaluation of this distributed refinement mechanism, including synchronization overhead and adaptation under heterogeneous network conditions, are left for future work.

\vspace{-0.15in}
\subsection{RL-Based Management Model}
\label{sec:scheduling}
The RL-based management system, illustrated in Fig. \ref{figure_2}, consists of three functional components: a resource allocation mechanism, an intelligent management controller, and a state compression module. At each transmission time interval (TTI), radio resources are allocated across active users while accounting for service-level fairness constraints. The intelligent controller adjusts parameters based on the network state, including compressed CQI information, user throughput statistics, and system load, to adapt the throughput-fairness trade-off. Due to the continuous and high-dimensional nature of the control problem, RL-based neural networks are employed to approximate optimal management policies. This work evaluates several RL algorithms \cite{comsa_thesis_2014}, each employing neural networks trained under different CQI compression configurations,  with the objective of maintaining the throughput-fairness operating point within the NGMN-defined feasible region over time.

The state dimension varies with the number of active users, requiring compression to reduce complexity. While basic QoS indicators can be summarized using statistical measures (e.g., mean and standard deviation), CQI reports form bandwidth-dependent vectors whose dimensionality scales with system bandwidth and user count. This paper proposes a CQI compression mechanism that extracts representative patterns from high-dimensional measurements, providing a compact, fixed-dimensional state representation for RL-based management.

\begin{figure*}[t]
\centering
\includegraphics[width=13cm]{./Figures/2_scar_scheduling_architecture.jpg}
\caption{RL-Based Management System.}
\centering
\label{figure_2}
\vspace{-0.25in}
\end{figure*}

\vspace{-0.00in}
\subsubsection{Fairness-Oriented Control}
We define $\mathcal{I}_t = \{1, 2, .., I_t\}$ as the set of active User Equipments (UEs) at TTI $t$, where $I_t$ denotes the number of users. Each user $i \in \mathcal{I}_t$ generates traffic to be managed under fairness constraints. The total system bandwidth is divided into $J$ Resource Blocks (RBs), represented by $\mathcal{J} = \{1, 2, ..., J\}$. Let $\mathbf{\Gamma}[t] = [\Gamma_1, \Gamma_2, \dots, \Gamma_{I_t}]$ denote the vector of average user throughputs at time $t$, where $\Gamma_i$ is the throughput of user $i$. 
The average throughput of user $i$ is updated at each TTI using an
exponential moving filter:
\vspace{-0.05in}
\begin{equation}
\label{eq:01}
\Gamma_i[t] = \lambda_{\mathrm{f}}\Gamma_i[t-1]
+ (1-\lambda_{\mathrm{f}}) \gamma_i[t]
\tag{1}
\vspace{-0.05in}
\end{equation}
where $\gamma_i[t]$ is the instantaneous user throughput and
$\lambda_{\mathrm{f}} \in [0,1)$ is the forgetting factor controlling
the influence of historical throughput measurements. 

The scheduler allocates RBs from $\mathcal{J}$ to users in $\mathcal{I}_t$, while the management controller adjusts its scheduling parameters to balance throughput and fairness. The NGMN fairness criterion \cite{ngmn_fairness} requires the CDF of $\Gamma_i[t]$ to lie to the right of a predefined reference line (see Fig.~\ref{figure_2}). To adapt to dynamic network conditions, the controller updates the scheduling parameters at each TTI by solving the following optimization problem:
\vspace{-0.05in}
\begin{equation}
\label{eq:02}
\begin{split}
\underset{a}{\max} \sum\nolimits_{i} \sum\nolimits_{j} a_{i,j}[t]  \cdot \frac{[\gamma(x_{i,j})]^{\beta_t}}{\Gamma_i^{\alpha_t}},  \qquad\qquad\qquad\qquad\qquad\quad   \\
s.t. \qquad\qquad\qquad\qquad\qquad\qquad\qquad\qquad\qquad\qquad\quad\quad\:\:\:\:\:\:\: \nonumber \\
\end{split}
\tag{2} 
\end{equation}

\begin{subequations}
\vspace{-0.25in}
\begin{align}
\label{eq:02a}
\sum\nolimits_{i} a_{i,j}[t] \leq 1, \quad j=1,2,...,J, \qquad\qquad\qquad\:\:\:\: \tag{2.a} \\
\label{eq:02b}
a_{i,j}[t] \in \{0,1\}, \quad \forall i \in \mathcal{I}_t, \forall j \in \mathcal{J}, \qquad\qquad\quad\:\:\:\: \tag{2.b} \\
\label{eq:02c}
\:\: \Upsilon_i(\Gamma_{i}, \mathbf{\Gamma}) \leq \Upsilon_{i}^{R}, \quad i=1,2,...,I_t. \qquad\qquad\qquad\: \tag{2.c} 
\end{align}
\vspace{-0.25in}
\end{subequations}

The optimization problem in (\ref{eq:02}) models the RB allocation process, where $a_{i,j} \in \{0,1\}$ is a binary decision variable indicating whether RB $j \in \mathcal{J}$ is assigned to UE $i \in \mathcal{I}_t$. The achievable rate $\gamma_{i,j}$ depends on the CQI value $x_{i,j} \in \{1, 2, \dots, N\}$, which determines the maximum number of bits transmittable on RB $j$ for user $i$. The optimization objective is controlled by parameters $\{\alpha_t, \beta_t\} \in [-1, 1]$, which balance system throughput and fairness:

\begin{itemize}
\vspace{-0.05in}
\item If $\beta_t > \alpha_t$, the management controller favors users with higher CQI values to maximize throughput.
\item If $\alpha_t \geq \beta_t$, it prioritizes users with lower average throughput to promote fairness.
\vspace{-0.05in}
\end{itemize}

Constraints (\ref{eq:02a}) ensure each RB is assigned to at most one user, while constraints (\ref{eq:02b}) reflect the combinatorial nature of the allocation. Constraints (\ref{eq:02c}) enforce the NGMN fairness criterion, where $\Upsilon_i(\Gamma_i, \mathbf{\Gamma})$ is the normalized CDF of user $i$’s throughput, and $\Upsilon_i^R$ is the required minimum value defined by the reference line shown in Fig. \ref{figure_2}.

The parameters $\alpha_t$ and $\beta_t$ constitute the control variables of the generalized proportional-fair (GPF) scheduling metric used in this work. While alternative scheduler parametrizations are possible, the adopted formulation captures the trade-off between throughput maximization and fairness through two directly interpretable quantities: $\beta_t$ favors users with better channel conditions, while $\alpha_t$ favors users with lower historical throughput. Consequently, adapting $[\alpha_t,\beta_t]$ enables the scheduler to steer operation toward the NGMN fairness region.

The solution to (\ref{eq:02}) aims to allocate each RB $j \in \mathcal{J}$ to the user $i \in \mathcal{I}_t$ that maximizes the ratio $[\gamma(x_{i,j})]^{\beta_t}/\Gamma_i^{\alpha_t}$, while satisfying the NGMN fairness constraint in (\ref{eq:02c}). However, due to varying channel conditions and user mobility, fixed parameter settings $\alpha_t$ and $\beta_t$ may lead to suboptimal outcomes, either unfair resource allocation (when $\beta_t > \alpha_t$) or overly fair allocations that degrade system throughput (when $\alpha_t > \beta_t$), as illustrated in Fig. \ref{figure_2}. 
To address this, the intelligent management controller must dynamically learn the optimal parameterization $[\alpha_t, \beta_t]$ at each TTI to keep the throughput CDF $\Upsilon_i$ as close as possible to the NGMN requirement line $\Upsilon_i^R$ (solid oblique line in Fig. \ref{figure_2}), while remaining above the defined lower threshold $\Upsilon_i^L$ (dotted oblique line in Fig. \ref{figure_2}). A solution is considered feasible when it satisfies the condition:
\vspace{-0.1in}
\begin{equation}
\label{eq:03}
\Upsilon_i^L \leq \Upsilon_i \leq \Upsilon_i^R.
\tag{3}
\vspace{-0.1in}
\end{equation}

The NGMN confidence interval is defined as the vertical distance between the reference fairness requirement and the lower admissible fairness boundary:
\vspace{-0.05in}
\begin{equation}
\label{eq:ngmn_confidence}
\Delta^{\mathrm{NGMN}} = \Upsilon_i^R - \Upsilon_i^L,
\tag{4}
\vspace{-0.05in}
\end{equation}
where $\Delta^{\mathrm{NGMN}}$ denotes the maximum admissible deviation from the NGMN reference line. Since the reference and lower fairness boundaries are parallel, this interval is identical for all active users at each TTI. For each user, the instantaneous distance from the NGMN reference line is computed as:
\vspace{-0.05in}
\begin{equation}
\label{eq:ngmn_distance}
\Delta_i[t] = \Upsilon_i^R - \Upsilon_i,
\tag{5}
\vspace{-0.05in}
\end{equation}
where $\Delta_i[t]$ measures the vertical distance between the required NGMN value and the achieved normalized CDF value for user $i$ at TTI $t$. 

\begin{figure*}[t]
\centering
\includegraphics[width=14cm]{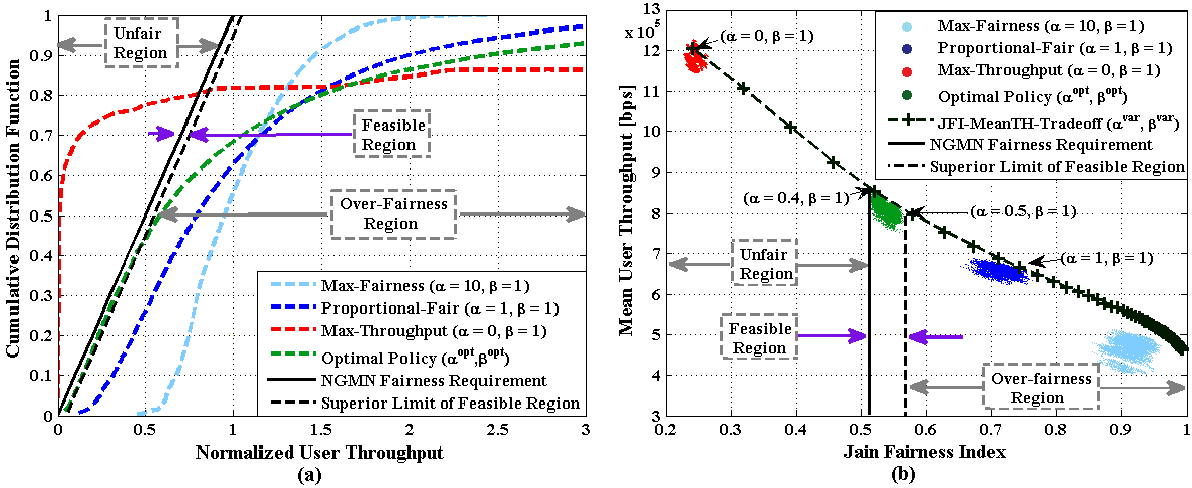}
\caption{Trade-off concept of system throughput and user fairness: (a) qualitative evaluation; (b) quantitative evaluation.}
\centering
\label{figure_2_1}
\vspace{-0.25in}
\end{figure*}

At each TTI $t$, the scheduler's operating region is determined from the deviations $\Delta_i[t]$ of active users within the normalized throughput CDF interval $[0.3,0.7]$, where compliance with the NGMN fairness requirement is evaluated. The aggregate deviation $\Delta_t$ is defined as the minimum deviation among these users, capturing the largest violation of the NGMN reference line when negative, or the smallest positive distance otherwise:
\vspace{-0.05in}
\begin{equation}
\label{eq:ngmn_margin}
\Delta_t = \min_{i \in \mathcal{I}_t^{*}} \Delta_i[t],
\quad
\mathcal{I}_t^{*} =
\{i \in \mathcal{I}_t : 0.3 \leq \Upsilon_i \leq 0.7\}.
\tag{6}
\vspace{-0.05in}
\end{equation}
Accordingly, the scheduler is classified as unfair ($\mathcal{UF}$) when $\Delta_t<0$, feasible ($\mathcal{FA}$) when:
\vspace{-0.05in}
\begin{equation}
\label{eq:ngmn_feasibility_condition}
0 \leq \Delta_t \leq \Delta^{\mathrm{NGMN}},
\tag{7}
\vspace{-0.05in}
\end{equation}
and overfair ($\mathcal{OF}$) when $\Delta_t>\Delta^{\mathrm{NGMN}}$. This
determines the reward type assigned to the RL controller at each TTI.

\subsubsection{Fairness Adaptation Mechanism} aims to maximize the timespent within the NGMN-defined feasible region (\ref{eq:ngmn_feasibility_condition}) while minimizing periods of unfair resource allocation. Through RL, the management controller learns to map each state $\mathbf{s}[t] \in \mathcal{S}$ to suitable scheduling fairness parameters $[\alpha_t, \beta_t]$:
\vspace{-0.05in}
\begin{equation}
\label{eq:08}
\begin{split}
\begin{cases}
\alpha_t = \alpha_{t-1} + \delta_{\alpha}, \\
\beta_t = \beta_{t-1} + \delta_{\beta},
\end{cases}
\end{split}
\tag{8}
\vspace{-0.15in}
\end{equation} 
where $[\delta_{\alpha}, \delta_{\beta}]$ denotes the fairness-adjustment action selected by the controller at TTI $t$, representing the increments applied to the fairness parameters $\alpha_{t-1}$ and $\beta_{t-1}$. Neural networks are used to approximate the optimal parameterization due to the continuous and multidimensional nature of the state space.

\subsubsection{Reward Formulation} 
To design the reward function, Fig.~\ref{figure_2_1} presents a qualitative and quantitative fairness evaluation for a static scenario with $I_t=60$ users uniformly distributed within a 1 km cell. Figure~\ref{figure_2_1}(a) shows throughput CDFs for three static scheduling configurations derived from the GPF rule: Maximum Fairness (MaxF, $\alpha=10$, $\beta=1$), Proportional Fair (PF, $\alpha=1$, $\beta=1$), and Maximum Throughput (MaxT, $\alpha=0$, $\beta=1$).
None consistently satisfy the NGMN feasible region, motivating RL-based adaptation of fairness parameters $[\alpha_t,\beta_t]$. Since the impact of a parameter update $[\delta_{\alpha},\delta_{\beta}]$ on the future CDF cannot be directly predicted, the reward combines qualitative NGMN fairness evaluation with the quantitative metrics shown in Fig.~\ref{figure_2_1}(b). At TTI $t$, the controller observes state $\mathbf{s}$, selects $[\delta_{\alpha},\delta_{\beta}]$, and updates the scheduling parameters $[\alpha_t,\beta_t]$. The scheduler then allocates resources using these parameters. At TTI $t+1$, the resulting state $\mathbf{s}'$ is classified into three fairness regions based on (\ref{eq:ngmn_margin}): unfair ($\mathcal{UF}$), feasible ($\mathcal{FA}$), or overfair ($\mathcal{OF}$). The reward $\rho_{t+1}$ is assigned according to the resulting fairness region:
\vspace{-0.05in}
\begin{equation}
\label{eq:09}
\begin{split}
\rho_{t+1}(\mathbf{s}, [\alpha_t, \beta_t])= \begin{cases}
\rho_{u}(\mathbf{s}, [\alpha_t, \beta_t]), & \textit{if $\mathbf{s}^{\prime} \in \mathcal{UF}$}, \\
1, & \textit{if $\mathbf{s}^{\prime} \in \mathcal{FA}$}, \\
\rho_{o}(\mathbf{s}, [\alpha_t, \beta_t]), & \textit{if $\mathbf{s}^{\prime} \in \mathcal{OF}$},
\end{cases}
\end{split}
\tag{9}
\vspace{-0.05in}
\end{equation}
where $\rho_u$ and $\rho_o$ denote the penalty functions associated with unfair and overfair operation, respectively.

When operating in the unfair region, the reward function encourages actions that move the throughput CDF toward the feasible region and penalizes those that increase unfairness. As seen in Fig. 3, increasing the scheduler’s emphasis on fairness improves compliance with the NGMN fairness criterion, while prioritizing throughput has the opposite effect. The reward magnitude depends on the current scheduling parameters and the selected adjustments $[\delta_{\alpha},\delta_{\beta}]$, as defined in (\ref{eq:10}).
\vspace{-0.05in}
\begin{equation}
\label{eq:10}
\begin{split}
\rho_{u} = \begin{cases}
\delta_{\alpha}, & \textit{if $\alpha_t \geq \beta_t$, $\delta_{\alpha} > 0$}, \\
-\delta_{\alpha}, & \textit{if $\alpha_t \geq \beta_t$, $\delta_{\alpha} \leq 0$}, \\
-1, & \textit{if $\alpha_t < \beta_t$, $\delta_{\alpha} \leq 0$,  $\delta_{\beta} \geq 0$}, \\
-0.5 \cdot (1 + |\delta_{\beta}|), & \textit{if $\alpha_t < \beta_t$, $\delta_{\alpha} \leq 0$,  $\delta_{\beta}  < 0$}, \\
-0.5 \cdot (|\delta_{\alpha}| + 1), & \textit{if $\alpha_t < \beta_t$, $\delta_{\alpha} > 0$,  $\delta_{\beta} \geq 0$}, \\
0.5 \cdot (|\delta_{\alpha}| + |\delta_{\beta}|), & \textit{if $\alpha_t < \beta_t$, $\delta_{\alpha} > 0$,  $\delta_{\beta} < 0$}.
\end{cases}
\end{split}
\tag{10}
\vspace{-0.15in}
\end{equation}
\vspace{-0.1in}

When operating in the overfair region, the reward function incentivizes actions that move the throughput CDF toward the feasible fairness region while avoiding unnecessary throughput degradation. In Fig.~\ref{figure_2_1}, reducing fairness emphasis relative to throughput increases throughput and shifts the operating point toward the feasible region. Actions that promote this transition are rewarded, while those that reinforce overfair operation are penalized. Similarly, the reward value depends on the current scheduler configuration and the selected fairness adjustments $[\delta_{\alpha},\delta_{\beta}]$. The resulting reward function is given in (\ref{eq:11}).
\vspace{-0.05in}
\begin{equation}
\label{eq:11}
\begin{split}
\rho_{o} = \begin{cases}
-1, & \textit{if $\alpha_t \geq \beta_t$, $\delta_{\alpha} \geq 0$,  $\delta_{\beta} \leq 0$}, \\
-0.5 \cdot (1 + |\delta_{\beta}|), & \textit{if $\alpha_t \geq \beta_t$, $\delta_{\alpha} \geq 0$,  $\delta_{\beta}  > 0$}, \\
-0.5 \cdot (|\delta_{\alpha}| + 1), & \textit{if $\alpha_t \geq \beta_t$, $\delta_{\alpha} < 0$,  $\delta_{\beta} \leq 0$}, \\
0.5 \cdot (|\delta_{\alpha}| + |\delta_{\beta}|), & \textit{if $\alpha_t \geq \beta_t$, $\delta_{\alpha} < 0$,  $\delta_{\beta} > 0$}, \\
\delta_{\beta}, & \textit{if $\alpha_t < \beta_t$, $\delta_{\beta} > 0$}, \\
-\delta_{\beta}, & \textit{if $\alpha_t < \beta_t$, $\delta_{\beta} \leq 0$}.
\end{cases}
\end{split}
\tag{11}
\vspace{-0.15in}
\end{equation}
\vspace{-0.15in}

\subsubsection{Controller's State}
At each TTI $t$, the RL controller observes a variable-dimensional state vector:
\vspace{-0.05in}
\begin{equation}
\label{eq:original_state}
\mathbf{s}[t]=[\alpha_{t-1},\beta_{t-1},\mathbf{x}[t],
\mathbf{\Gamma}[t], \Delta_t, \digamma_t, I_t],
\tag{12}
\vspace{-0.05in}
\end{equation}
where $\mathbf{x}[t]=[\mathbf{x}_1[t],\ldots,\mathbf{x}_{I_t}[t]]$ contains user CQI vectors, with $\mathbf{x}_i[t]=[x_{i,1}[t],\ldots,x_{i,J}[t]]$, and $\mathbf{\Gamma}[t]$ contains user throughputs. Distance $\Delta_t$ denotes the aggregate NGMN deviation determined based on (\ref{eq:ngmn_margin}), while $\digamma_t\in\{-1,0,1\}$ indicates unfair, overfair, and feasible operation, respectively. Under its original representation, the state dimension varies with the number of active users $I_t$ and system bandwidth.

To provide a compact state representation for RL-based management and support operation within the NGMN-defined feasible region, SCAR transforms the variable-dimensional state vector $\mathbf{s}$ into a fixed-dimensional representation. While the user throughput vector $\mathbf{\Gamma}[t]$ can be compressed using its mean and standard deviation \cite{comsa_thesis_2014}, CQI compression is more challenging due to its high dimensionality and variability. SCAR addresses this through a learning-based CQI compression mechanism that extracts compact features for RL-based resource management under service-level fairness constraints.

\vspace{-0.25in}
\subsection{SAST-Based Optimization} 
\label{sec:SAST}
\vspace{-0.05in}
To enhance the performance of the SCAR framework in both CQI clustering and classification, we adopt an optimization technique based on Simulated Annealing with Stochastic Tunneling (SAST), inspired by the adaptive stochastic tunneling framework proposed in \cite{5621041}. This meta-heuristic method is well-suited for exploring large, non-convex search spaces and finding globally optimal solutions under uncertainty.

In the context of SCAR, the SAST algorithm is applied to two key problems: (i) the selection of optimal CQI cluster centers during offline K-means clustering, and (ii) the tuning of RBFN weights in the online classification phase. Let $\mathcal{B}$ denote the current state (e.g., a set of CQI cluster centers or RBFN weights), and $\mathcal{A}$ a candidate neighboring solution. The acceptance probability of transitioning from $\mathcal{B}$ to $\mathcal{A}$ is\cite{5621041}:
\vspace{-0.05in}
\begin{equation}
\label{eq:13}
\mathbb{P}(\mathcal{A} | \mathcal{B}) = \min \left[1, \mathbb{F}(\mathcal{A}, \mathcal{B}, \mathbf{T}) \right],
\tag{13}
\vspace{-0.05in}
\end{equation}
where $\mathbf{T}$ is a global temperature parameter, and $\mathbb{F}(\cdot)$ is the acceptance function\cite{5621041}:
\vspace{-0.05in}
\begin{equation}
\label{eq:14}
\mathbb{F}(\mathcal{A}, \mathcal{B}) = \exp \left[ - \frac{F(\mathcal{A}) - F(\mathcal{B})}{\mathbf{T}} \right],
\tag{14}
\vspace{-0.05in}
\end{equation}
with $F(\cdot)$ being the stochastic tunneling function that normalizes the 
energy relative to the best solution found so far \cite{5621041}:
\vspace{-0.1in}
\begin{equation}
\label{eq:15}
F(\mathcal{X}) = 1 - exp \Bigg[ - \frac{f(\mathcal{X}) - f^{*}}{\omega} \Bigg],
\tag{15}
\vspace{-0.05in}
\end{equation}
where $\mathcal{X} \in \{\mathcal{A}, \mathcal{B}\}$, $f(\mathcal{X})$ is the energy (cost) of state $\mathcal{X}$, $f^{*}$ is the minimum energy observed so far, and $\omega$ is the tunneling parameter. In the CQI clustering problem, $f(\mathcal{X})$ represents the Euclidian distance between CQI centers and other CQI points, while in RBFN training, it may denote the classification loss. The SAST mechanism enables escape from local minima by comparing energy values against $f^*$ and accepting worse states with a probability that decreases as the temperature $\mathbf{T}$ lowers. The initial temperature is computed as \cite{5621041}:
\vspace{-0.1in}
\begin{equation}
\label{eq:16}
\mathbf{T}_0 = - \sum\nolimits_{l=1}^{L} [F_l(\mathcal{A}) -  F_l(\mathcal{B})] \big/ [ L \cdot ln(\mathbb{P}_0)],
\tag{16}
\vspace{-0.05in}
\end{equation}
where $L$ is the number of sampling iterations and $\mathbb{P}_0$ is the initial acceptance probability. The temperature is then gradually reduced at each iteration based on:
\vspace{-0.1in}
\begin{equation}
\label{eq:17}
\mathbf{T}_{new} = \mathbf{T}_{old} \cdot R_{T}, 
\tag{17}
\vspace{-0.1in}
\end{equation}
where $R_T \in [0,1]$ is the cooling rate.

By applying SAST to the clustering and RBFN optimization stages, SCAR improves the representation of CQI states while retaining computationally efficient online state classification. The resulting compact CQI representation is subsequently used as input to the RL-based RRM controller operating under throughput and service-level fairness objectives.

\vspace{-0.15in}
\section{Proposed CQI Compression Mechanism} \label{sec:compression}
As shown in Fig. \ref{figure_2}, the proposed CQI compression mechanism operates in three stages. First, a pre-processing stage removes system bandwidth dependency from CQI reports. Next, the classification stage uses a two-step learning approach: K-means clustering (unsupervised) identifies structural patterns in CQI data, while a RBF network (supervised) classifies new CQI inputs into these learned clusters. Finally, the statistical stage extracts key features from the classified CQI space to support computationally efficient edge-assisted operation.

\vspace{-0.1in}
\subsection{Preprocessing Stage}
\label{sec:preprocessing}
The preprocessing stage transforms each CQI report $\mathbf{x}_i = [x_{i,1}, \dots, x_{i,J}]$ from bandwidth-dependent vectors into normalized histograms independent of system bandwidth. This is done by computing the frequency distribution of CQI values:
\vspace{-0.05in}
\begin{equation}
\label{eq:18}
\mathbf{y}_{i} = [y_{i,1}, y_{i,2}, ..., y_{i,N}],
\tag{18}
\vspace{-0.05in}
\end{equation}
where $y_{i,n}$ 
is the normalized count of CQI value $n$ for user $i$:
\vspace{-0.2in}
\begin{align}
\label{eq:19a}
y_{i,n}=\frac{1}{J \cdot n} \sum\nolimits_{\substack{j=1 \\ x_{i,j}=n}}^{J} x_{i,j}, \forall i \in \mathcal{I}_t, n=\{1,...,N\}, \tag{19.a}\\
\vspace{-0.05in}
\label{eq:19b}
\sum\nolimits_{n=1}^{N} y_{i,n}=1, \forall i \in \mathcal{I}_t. \qquad\qquad\qquad\qquad\qquad\:\:\:
\vspace{-0.05in}
\tag{19.b}
\end{align}

Although this reduces dimensionality from $J$ to $N$, many values in $\mathbf{y}_i$ remain insignificant. To further reduce complexity, the top-$M$ most significant entries are selected:
\vspace{-0.05in}
\begin{align}
\label{eq:20a}
\mathcal{M}_i &= \{y_{i,n,1}, ..., y_{i,n,M}\}, \tag{20.a} \\
\label{eq:20b}
\mathcal{R}_i &= \{y_{i,n,M+1}, ..., y_{i,n,N}\}, \tag{20.b}
\vspace{-0.1in}
\end{align}
where the elements in $\mathbf{y}_i$ are sorted in descending order:
\vspace{-0.05in}
\begin{equation}
\label{eq:21}
y_{i,n,1} > y_{i,n,2} > \cdots > y_{i,n,N}. \tag{21}
\end{equation}

Residual values $\mathcal{R}i$ are redistributed to the nearest top-$M$ values from $\mathcal{M}_i$ to preserve normalization:
\vspace{-0.05in}
\begin{equation}
\label{eq:22}
y_{i,n, p} = 
    \begin{cases} 
        y_{i, n, m} + \sum\nolimits_{r_m}^{R_m} y_{i,n,r_m}, &  m \leq M  \\
        0, & otherwise,
   \end{cases}
\tag{22}
\vspace{-0.05in}
\end{equation}
resulting in a bandwidth-independent preprocessed vector:
\vspace{-0.05in}
\begin{equation}
\label{eq:23}
\mathbf{y}_{i,M} = [y_{i,n,1}, y_{i,n,2},..., y_{i,n,M}, 0,...,0],
\tag{23}
\vspace{-0.05in}
\end{equation}
with elements reordered according to the original CQI index $n$. These compact vectors are then used in the classification stage to group similar CQI patterns. A two-step learning process is applied: \textit{a)} unsupervised learning (K-means) identifies $K$ representative CQI patterns; and \textit{b)} supervised learning (RBFN) classifies each $\mathbf{y}_{i,M}$ to one of the $K$ classes. This yields a classification vector at each TTI:
\begin{equation}
\label{eq:24}
\mathbf{v}_M[t] = [v_{M,1}, v_{M,2}, ..., v_{M,K}], \tag{24}
\end{equation}
where $v_{M,k}$ denotes the number of users whose CQI patterns belong to class $k$. The joint parameterization of $\{M, K\}$ determines the level of CQI compression and the resulting state representation provided to the RL-based RRM controller.

\vspace{-0.15in}
\subsection{Unsupervised Machine Learning}
\label{sec:clustering}
In SCAR, unsupervised learning is used to identify representative clusters from pre-processed CQI data. Let $\mathcal{U}_M = \{\mathbf{y}_{u,M} \mid u = 1, 2, ..., U_M\}$ be the set of $U_M$ distinct top-$M$ pre-processed CQI vectors. The goal is to compute a set of $K_M$ cluster centers $\mathcal{K}_M = \{\mathbf{c}_k \mid k = 1, 2, ..., K_M \}$, where each center $\mathbf{c}_k = [c_{1,k}, c_{2,k},...,c_{N,k}]$ lies in the $N$-dimensional space of pre-processed CQI reports. The clustering objective is to minimize the average distortion, defined as the mean squared Euclidean distance between each data point $\mathbf{y}_{u,M}$ and its assigned cluster centroid $\mathbf{c}_{k(u)}$:
\vspace{-0.05in}
\begin{equation}
\label{eq:25}
d(\mathcal{K}_M) = 1 / U_M \sum\nolimits_{u=1}^{U_M} \left \| \mathbf{y}_{u,M} - \mathbf{c}_{k(u)}\right\|^2,
\tag{25}
\vspace{-0.05in}
\end{equation}
where $\mathbf{c}_{k(u)}$ is the closest center to $\mathbf{y}_{u,M}$. This minimization can be achieved using classical clustering techniques such as K-means or more robust alternatives like Swap heuristics \cite{mount_sa}, iteratively refining the center set $\mathcal{K}_M^{(z)}$ over $z = 1, 2, ..., Z_C$. These methods serve as the foundation for the proposed SAST-based optimization strategy presented in Sub-section \ref{sec:SAST}.4.

\noindent \subsubsection{Iterative K-means Algorithm}
At each iteration $z$, K-means performs:

\noindent \textit{a)} Assignment: each data point $\mathbf{y}_{u,M} \in \mathcal{U}_M$ is assigned to its nearest center $\mathbf{c}_k^{(z-1)} \in \mathcal{K}^{(z-1)}$, forming neighborhoods $\mathcal{N}_k$;

\noindent  \textit{b)} Update: new centers are computed as the mean of their assigned points:
\vspace{-0.1in}
\begin{equation}
\label{eq:26}
\bar{\mathbf{c}}_{k}^{(z-1)} = 1/|\mathcal{N}_{k}| \cdot \sum\nolimits_{u}^{|\mathcal{N}_{k}|} \mathbf{y}_{u,M},
\tag{26}
\vspace{-0.1in}
\end{equation}
and set as $\mathbf{c}_k^{(z)} = \bar{\mathbf{c}}_k^{(z)}$;

\noindent  \textit{c)} Reassignment: data points are reassigned to the updated nearest centers. This loop continues for $Z_C$ iterations or until convergence. Two main challenges arise: (i) neighborhood search becomes computationally expensive as $U_M$ increases, which can be mitigated using KN-trees for faster lookup \cite{mount_kd_tree}; (ii) convergence to local minima, inherent in K-means, can be improved through enhanced initialization of centers and convergence criteria.

\subsubsection{KN-Tree Accelerated Clustering}
To reduce the computational complexity of K-means clustering for large CQI datasets, a kd-tree-based acceleration method is adopted \cite{mount_kd_tree}. The pre-processed CQI points in $\mathcal{U}_M$ are recursively partitioned into hierarchical cells, enabling efficient nearest-center assignments. During clustering, candidate centers are filtered using geometric distance bounds, eliminating centers that cannot be nearest to any point within a cell. When only one candidate remains, all points in that cell are assigned directly to the corresponding center, avoiding unnecessary distance calculations. The resulting neighborhoods $\mathcal{N}_k^{(z)}$ are used to update the centers according to (\ref{eq:26}).

\subsubsection{Center Refinement and Swap-Based Initialization}
To improve clustering quality, the K-means procedure is executed through multiple runs (epochs). Each run starts from an initial center set $\mathcal{K}_M^{0}$ and iteratively updates the centers until convergence. During the process, the best center configuration found so far is tracked as
\vspace{-0.05in}
\begin{equation}
\label{eq:27}
\mathcal{K}^{*}_M =
\begin{cases}
\mathcal{K}^{(z)}_M, & d^{(z)} < d^{*},\\
\mathcal{K}^{*}_M, & \text{otherwise},
\end{cases}
\tag{27}
\vspace{-0.05in}
\end{equation}
where $d^{*}$ denotes the minimum distortion observed up to the current iteration. At the end of a run, the resulting center configuration is used as the initialization of the next run. While this refinement strategy progressively improves the solution quality, the search remains local and may converge to suboptimal minima. To increase exploration and reduce the likelihood of local trapping, a Swap heuristic is incorporated. Instead of initializing a new run from the previous center configuration, centers are randomly selected from the dataset $\mathcal{U}_M$, i.e., $\mathcal{K}_M^{(z)} \subset \mathcal{U}_M$. This randomized re-initialization allows the search to explore different regions of the solution space and increases the probability of discovering lower-distortion clustering configurations.

\subsubsection{Proposed SAST Meta-Heuristic Clustering}
To balance the exploration capabilities of random center selection and the precision of KN-tree-based clustering, we propose a meta-heuristic algorithm that dynamically switches between both strategies. Unlike conventional KN-tree clustering, which discards non-improving center sets, our approach evaluates whether such sets might still guide the search toward better solutions. Similarly, while random center selection can escape local minima, it often yields high distortion. By alternating between methods and making probabilistic decisions on whether to retain or switch center sets, our framework aims to efficiently discover better clustering configurations.

At each iteration $z$, given the previous set of centers $\mathcal{K}_M^{(z-1)}$, we solve the following optimization problem to compute the new set $\mathcal{K}_M^{(z)}$, as follows:
\begin{equation}
\label{eq:28}
\min_{\boldsymbol{\psi}, \boldsymbol{\zeta}, \boldsymbol{\varphi}} 
\sum_{u=1}^{U_M} \sum_{k=1}^{K_M} \psi_{u,k} \cdot \left\| 
\mathbf{y}_u - \sum_{l=1}^{2} \xi_{k,l} \sum_{r=1}^{U_M} 
\zeta_{k,l} \cdot \varphi_{k,r} \cdot \mathbf{y}_r \right\|^2
\tag{28}
\end{equation}
\noindent where: $\psi_{u,k} \in \{0,1\}$ assigns data point $u$ to center $k$; $\zeta_{k,l} \in \{0,1\}$ selects the clustering method $l \in \{1\ (\text{random initialization}),\ 2\ (\text{KN-tree computation})\}$ for center $k$; $\varphi_{k,r} \in \{0,1\}$ indicates whether point $r$ contributes to computing center $k$; and $\xi_{k,l} = 
    \begin{cases}
    1, & \text{if } l = 1 \\
    1 / |\mathcal{N}_k |, & \text{if } l = 2
    \end{cases}$ is a normalization factor depending on the clustering method.
This objective balances assignment accuracy based on average distortion, method flexibility, and adaptive center selection, forming the basis of our SAST-driven clustering procedure.

The proposed SAST-based clustering leverages temperature-driven probabilistic transitions between two strategies: random center selection and KN-tree-based refinement. Let $\mathcal{K}_{M,1}^{(z)}$ denote the set of centers at iteration $z$ obtained using random selection. At each iteration, the state pair $(\mathcal{A}, \mathcal{B}) = (\mathcal{K}_{M,1}^{(z)}, \mathcal{K}_{M,1}^{(z-1)})$ is evaluated. If the distortion $d(\mathcal{K}_{M,1}^{(z)}) \leq d(\mathcal{K}_{M,1}^{(z-1)})$, the random strategy is retained. Otherwise, SAST computes the acceptance probability $\mathbb{P}(\mathcal{A} | \mathcal{B})$, which decreases as temperature cools, favoring a switch to KN-tree-based clustering. In a similar manner, when a run terminates with $d(\mathcal{K}^{(z)}_M) > d(\mathcal{K}^{0}_M)$, the state pair $(\mathcal{A}, \mathcal{B}) = (\mathcal{K}^{(z)}_M, \mathcal{K}^{0}_M)$ is evaluated to decide whether to carry over the current centers or reinitialize them.

\begin{table}[t]
\centering
\scriptsize
\line(1,0){245}
\scriptsize 
\renewcommand{\caption}{\small}
\caption{\small \bf{Algorithm I: SAST-based Clustering}}
\line(1,0){245} 
\begin{algorithmic}[1]
\small
\State
{\bf while} \textit{($run_{new} = true$)} \& ($z \leq Z_C$) {\bf do} \\
{${}$\hspace{1em} \bf start} iteration $z$ \\
{${}$\hspace{2em} \bf if } \textit{(random = true)} \\
{${}$\hspace{3em} \bf randomize } one center $\forall \mathbf{c}_k \in \mathcal{K}^{(z-1)}_M$, $k=1,..,K_M$ \\
{${}$\hspace{2em} \bf else} iterated KN-tree \\
{${}$\hspace{3em} \bf determine} neighborhood $\mathcal{N}_k$ for each $\mathbf{c}_k \in \mathcal{K}^{(z-1)}_M$ \\ 
{${}$\hspace{3em}} based on the filtering approach \\
{${}$\hspace{3em} \bf calculate} centroid $\bar{\mathbf{c}}_k^{(z)}$ based on (\ref{eq:26}) and $\mathbf{c}_k^{(z)}=\bar{\mathbf{c}}_k^{(z)}$ \\
{${}$\hspace{2em} \bf end if} \\
{${}$\hspace{2em} \bf assign} each point $\mathbf{y}_{u,M} \in \mathcal{U}_M$ to closest $\mathbf{c}_k \in \mathcal{K}^{(z)}_M$ \\
{${}$\hspace{2em} \bf calculate} distortion $d(\mathcal{K}^{(z)}_M)$ based on (\ref{eq:25}) \\
{${}$\hspace{2em} \bf if} $(z < max_{iterations/run})$ {\bf or} ($\bar{d}^{(z)} > \bar{d}_{min}$) \\
{${}$\hspace{3em}} $run_{new} = false$ \\
{${}$\hspace{3em} \bf calculate} $\mathbb{P}(\mathcal{K}^{(z)}_{M,1} | \mathcal{K}^{(z-1)}_{M,1})$ based on (\ref{eq:13})-(\ref{eq:15}) \\
{${}$\hspace{3em} \bf decrease} temperature based on (\ref{eq:16})-(\ref{eq:17}) \\
{${}$\hspace{3em} \bf if} $[\mathbb{P}(\mathcal{K}^{(z)}_{M,1} | \mathcal{K}^{(z-1)}_{M,1}) > \mathbb{P}_{rand}]$: $random = true$ \\
{${}$\hspace{3em} \bf else:} $random = false$ \\
{${}$\hspace{2em} \bf else} \\
{${}$\hspace{3em}} $run_{new}=true$, $random=true$ \\
{${}$\hspace{3em} \bf calculate} $\mathbb{P}(\mathcal{K}^{(z)}_M | \mathcal{K}^{0}_M)$ based on (\ref{eq:13})-(\ref{eq:15}) \\
{${}$\hspace{3em} \bf decrease} temperature $\mathbf{T}_{old}$ based on (\ref{eq:16})-(\ref{eq:17}) \\
{${}$\hspace{3em} \bf if} $(d^{(z)} < d^{*})$ $d^{*} = d^{(z)}$ based on (\ref{eq:27}) \\
{${}$\hspace{3em} \bf if} $(\mathbb{P}(\mathcal{K}^{(z)}_M | \mathcal{K}^{0}_M) > \mathbb{P}_{rand})$ \\
{${}$\hspace{4em}} Initial set for the next run epoch: $\mathcal{K}^{0}_M=\mathcal{K}^{(z)}_M$ \\
{${}$\hspace{3em} \bf else} keep $\mathcal{K}^{0}_M$ in next run \\
{${}$\hspace{2em} \bf end if} \\
{${}$\hspace{1em} \bf end} iteration $z$ \\
{\bf end while}
\vspace{-0.1in} 
\end{algorithmic}
\line(1,0){245}
\vspace{-0.2in}  
\end{table}

Algorithm I presents the proposed SAST-based meta-heuristic approach designed to efficiently cluster large datasets $\mathcal{U}_M$ containing pre-processed CQI reports. Each execution begins with a random selection of initial centers, followed by the assignment of data points to their respective clusters. If neither the maximum number of iterations per run has been reached nor the minimum distortion value achieved, the algorithm evaluates whether to switch to the KN-tree-based clustering method, as indicated in lines (13)-(17). If the maximum number of iterations for a run is reached, the current iteration becomes the final one for that run, and the SAST method determines whether to store a new initial set of centers for the subsequent run (lines (19)-(25)). This process continues iteratively until the max number of runs is completed.

Upon completion, the best solution $\mathcal{K}^{*}_M$ is sorted according to CQI quality. Consider the baseline set $\mathcal{U}_1$ of top-$1$ CQI reports with $U_1 = N$, where $\mathbf{y}_{1,1} = [1, 0, \dots, 0]$ and $\mathbf{y}_{N,1} = [0, \dots, 0, 1]$ correspond to the worst and best CQI, respectively. For each center $\mathbf{c}_k \in \mathcal{K}^{*}_M$, define the nearest CQI index as follows:
\vspace{-0.1in}
\begin{equation}
\label{eq:29}
n_k = \arg\min_{n \in \{1,\dots,N\}} \left\| \mathbf{c}_k - \mathbf{y}_{n,1} \right\|^2.
\tag{29}
\vspace{-0.05in}
\end{equation}
Then, define the ordering permutation $\pi$ such that $n_{\pi(1)} < n_{\pi(2)} < \dots < n_{\pi(K_M)}$. The final ordered center set is:
\vspace{-0.05in}
\begin{equation}
\label{eq:30}
\mathcal{K}_M = \{ \mathbf{c}_{\pi(k)} \}_{k=1}^{K_M},
\tag{30}
\vspace{-0.05in}
\end{equation}
ensuring that $\mathbf{c}_{\pi(1)}$ and $\mathbf{c}_{\pi(K_M)}$ represent the clusters of worst and best average channel conditions, respectively.

\vspace{-0.15in}
\subsection{Supervised Machine Learning}
\label{sec:rbfn}
The role of supervised learning is to construct a prediction model that assigns each pre-processed CQI report $\mathbf{y}_{i,M}$, for user $i \in \mathcal{I}_t$, to a corresponding discrete pattern. Define the set of patterns as $\mathcal{P} = \{\mathbf{p}_1, \mathbf{p}_2, ..., \mathbf{p}_K\}$, where each pattern $\mathbf{p}_k = [p_{k,1}, p_{k,2}, ..., p_{k,O}]$ is a binary encoding of the cluster index $k$ derived from center $\mathbf{c}_k \in \mathcal{K}_M$, with $0$ values replaced by $-1$ such that $p_{k,o} \in \{-1, 1\}$ for $o=1,...,O$. Given the collection of pre-processed CQI vectors $\mathcal{U}_M$, the goal is to learn a mapping from the continuous input space to the finite label space $\mathcal{P}$, using the labeled dataset:
\begin{equation}
\label{eq:31}
\mathcal{V}=\{(\mathbf{y}_{u}, \mathbf{p}_{u}) | \mathbf{y}_{u} \in \mathcal{U}_M, \mathbf{p}_{u} \in \mathcal{P}, u=1,...,U_M\}.
\tag{31}
\end{equation}

Based on the labeled dataset $\mathcal{V}$, the goal is to learn a non-linear function $\Phi: \mathbb{R}^{N} \rightarrow [-1,1]^{O}$ that maps each pre-processed CQI vector $\mathbf{y}_{u,M}$ to its corresponding pattern $\mathbf{p}_u$:
\vspace{-0.05in}
\begin{equation}
\label{eq:32}
\Phi(\mathbf{y}_{u,M}) = \mathbf{p}_{u}, \quad u = 1, \dots, U_M.
\tag{32}
\vspace{-0.05in}
\end{equation}
This mapping is implemented using a RBF network:
\vspace{-0.05in}
\begin{equation}
\label{eq:33}
\Phi(\mathbf{y}_{u,M}) = \Phi_2 \left[ \mathbf{W} \cdot \Phi_1(\mathbf{y}_{u,M}^{T}) \right],
\tag{33}
\vspace{-0.05in}
\end{equation}
where $\mathbf{W} = \{w_{k,o}\}$ is the matrix connecting the $K$ hidden units to $O$ output nodes. The hidden layer output is the vector $\Phi_1 = [\phi_{u,1}^{(1)}, \dots, \phi_{u,K}^{(1)}]$, where each $\phi$ is a Gaussian RBF:
\vspace{-0.05in}
\begin{equation}
\label{eq:34}
\phi_{u,k}^{(1)} = \exp \left( -\frac{\|\mathbf{y}_{u,M} - \mathbf{c}_k\|^2}{2\sigma^2} \right),
\vspace{-0.05in}
\tag{34}
\end{equation}
with fixed $\sigma > 0$. Finally, $\Phi_2 = [\phi_1^{(2)}, \dots, \phi_O^{(2)}]$ applies a $\tanh(\cdot)$ activation function component-wise at the output layer.

Perfect interpolation in (\ref{eq:32}) is generally unattainable due to the impracticality of collecting the complete space of CQI vectors $\mathcal{U}_M$, as many combinations are never observed under realistic channel fading conditions. Therefore, the RBF-based mapping $\Phi$ is approximated as $\widetilde{\Phi}$. To improve generalization, we construct a new training set $\mathcal{T}$ using CQI reports $\mathbf{y}_{i,M} \in \mathcal{Y}_M$ gathered from network simulations:
\vspace{-0.05in}
\begin{equation}
\label{eq:35}
\mathcal{T} = \{(\mathbf{y}_{i,M}, \mathbf{p}_i) \mid \mathbf{y}_{i,M} \in \mathcal{Y}_M, \mathbf{p}_i \in \mathcal{P} \}.
\tag{35}
\vspace{-0.05in}
\end{equation}
For samples in $\mathcal{T} \setminus \mathcal{V}$, the model employs:
\vspace{-0.05in}
\begin{equation}
\label{eq:36}
\widetilde{\Phi}(\mathbf{y}_{i,M}) + E(\mathcal{W}, \mathbf{y}_{i,M}) = \mathbf{p}_i,
\tag{36}
\vspace{-0.05in}
\end{equation}
where $E$ denotes the approximation error as a function of weights from set $\mathcal{W}$. Integrating this training with a network simulator allows the RBF model to adaptively learn under realistic CQI variations and better generalize beyond the original labeled set $\mathcal{V}$.

Similarly, the RBF network is trained over $Z_R$ iterations grouped into epochs aligned with TTIs. At each iteration $z \in \{1,2,\dots,Z_R\}$, the goal is to minimize the mean square error:
\begin{equation}
\label{eq:37}
e^{(z)} = 1/O \cdot ||\mathbf{p}_{i} - \mathbf{y}_{i,M}^{(2)}||^{2} = 1/O \cdot \sum\nolimits_{o}^{O} ({p}_{i,o}-{y}_{i,o}^{(2)})^{2},
\tag{37}
\end{equation}
where $\mathbf{y}_{i,M}^{(2)} = \Phi_2(\cdot)$ is the RBF output vector. The best set of weights $\mathcal{W}^{*}$ is updated according to:
\vspace{-0.05in}
\begin{equation}
\label{eq:38}
\mathcal{W}^{*} = 
    \begin{cases} 
        \mathcal{W}^{(z)}, & \text{if } e^{(z)} < e^{*} \\
        \mathcal{W}^{*}, & \text{otherwise},
    \end{cases}
\tag{38}
\vspace{-0.05in}
\end{equation}
where $e^*$ is the lowest error observed so far. Each iteration includes: \textit{a)} forward pass of $\mathbf{y}_M$, \textit{b)} error computation and backpropagation, and \textit{c)} weight update to avoid local minima.

\subsubsection{Forward Propagation.}
The output $\mathbf{y}_M^{(2)}$ is computed from input $\mathbf{y}_M$ and weights $\mathcal{W}^{(z)}$. To avoid overfitting from relying solely on $\mathcal{T}$, we alternate between $\mathcal{T}$ and $\mathcal{V}$ using a SAST-driven management controller. Each run starts with $\mathcal{T}$ and switches to $\mathcal{V}$ based on energy states $\mathcal{A} = \mathcal{W}_1^{(z)}$ and $\mathcal{B} = \mathcal{W}_1^{(z-1)}$, with $f = e$, $f^* = e^*$. If $F(\mathcal{W}_1^{(z)}) \geq F(\mathcal{W}_1^{(z-1)})$, training continues with $\mathcal{T}$; otherwise, $\mathcal{W}_1^{(z)}$ is accepted with probability $\mathbb{P}(\mathcal{W}_1^{(z)} | \mathcal{W}_1^{(z-1)})$. If this is less than $\mathbb{P}_{rand}$, the switch to $\mathcal{V}$ occurs from iteration $z+1$.

\subsubsection{Error Back-Propagation.}
For each training sample from $\mathcal{T}$ or $\mathcal{V}$, the input $\mathbf{y}_M$ produces the output $\mathbf{y}^{(2)}_M$ via forward propagation. The error vector $\mathbf{e} = \mathbf{p} - \mathbf{y}^{(2)}_M$ is computed and back-propagated through the output layer using:
\begin{equation}
\label{eq:39}
\mathbf{e}^{(2)} = \bigtriangledown\Phi^{(2)} \circ \mathbf{e},
\tag{39}
\vspace{-0.05in}
\end{equation}
where $\bigtriangledown\Phi^{(2)}$ is the vector of derivatives of the activation functions, and $\circ$ denotes element-wise multiplication. Let $\mathbf{y}_M^{(1)}$ be the hidden layer output and $\mathbf{W}^{(z-1)}$ the previous weight matrix, the weights are updated by:
\vspace{-0.1in}
\begin{equation}
\label{eq:40}
\mathbf{W}^{(z)} = \mathbf{W}^{(z-1)} + \eta \cdot \mathbf{y}^{(1)T}_M \times \mathbf{e}^{(2)},
\tag{40}
\vspace{-0.1in}
\end{equation}
where $\eta$ is the learning rate.

\begin{table}[t]
\centering
\scriptsize
\line(1,0){245}
\scriptsize 
\renewcommand{\caption}{\small}
\caption{\small \bf{Algorithm II: SAST- based RBFN Training}}
\line(1,0){245} 
\begin{algorithmic}[1]
\small
\State
{\bf for} each run epoch (TTI $t$) \\
{\bf} $\mathcal{V} = false$ \\
{${}$\hspace{1em} \bf for} each iteration $z$ (user $i \in \mathcal{I}_t$) \\
{${}$\hspace{2em} \bf receive} CQI report $\mathbf{x}_i$ \\
{${}$\hspace{2em} \bf process} $\mathbf{x}_i$ to $\mathbf{y}_{i,M}$ based on (\ref{eq:18})-(\ref{eq:23}) \\
{${}$\hspace{2em} \bf if} ($\mathcal{V} = false$) pick $\mathbf{y}_{i,M} \in \mathcal{Y}_M$ from environment \\
{${}$\hspace{2em} \bf else} pick $\mathbf{y}_{u,M} \in \mathcal{U}_M$ $\forall u = 1,2,...,U_M$ \\
{${}$\hspace{2em} \bf determine} pattern $\mathbf{p} \in \mathcal{P}$ for the selected $\mathbf{y}_M$ \\
{${}$\hspace{2em} \bf propagate} $\mathbf{y}_M$ through RBFN based on (\ref{eq:32})-(\ref{eq:34}) \\
{${}$\hspace{2em} \bf calculate} error $e^{(z-1)}$ based on (\ref{eq:36}) and (\ref{eq:37}) \\
{${}$\hspace{2em} \bf back-propagate} error vector $\mathbf{e}^{(z-1)}$ based on (\ref{eq:39}) \\
{${}$\hspace{2em} \bf correct} weights $\mathcal{W}^{(z-1)}$ based on (\ref{eq:40}) \\
{${}$\hspace{2em} \bf calculate} error $e^{(z)}$ based on (\ref{eq:36}) and (\ref{eq:37}) \\
{${}$\hspace{2em} \bf determine} prob. $\mathbb{P}(\mathcal{W}^{(z)}_1 | \mathcal{W}^{(z-1)}_1)$ based on (\ref{eq:13})-(\ref{eq:15}) \\
{${}$\hspace{2em} \bf if} $[\mathbb{P}(\mathcal{W}^{(z)}_1 | \mathcal{W}^{(z-1)}_1) > \mathbb{P}_{rand}]$ {\bf then} $\mathcal{V} = false$ \\
{${}$\hspace{2em} \bf else} $\mathcal{V} = true$ \\
{${}$\hspace{2em} \bf if} iteration $z=I_t$ \\
{${}$\hspace{3em} \bf if} ${e}^{(z)} < e^0$ \\
{${}$\hspace{4em} \bf verify} in (\ref{eq:38}) if $\mathcal{W}^{*} = \mathcal{W}^{(z)}$ and set $\mathcal{W}^{0} = \mathcal{W}^{(z)}$ \\
{${}$\hspace{3em} \bf else} // non-better $\mathcal{W}^{(z)}$ acceptance \\
{${}$\hspace{4em} \bf determine} $\mathbb{P}(\mathcal{W}^{(z)} | \mathcal{W}^{0})$ based on (\ref{eq:13})-(\ref{eq:15}) \\
{${}$\hspace{4em} \bf if} $[\mathbb{P}(\mathcal{W}^{(z)} | \mathcal{W}^{0}) > \mathbb{P}_{rand}]$ {\bf then} $\mathcal{W}^{0} = \mathcal{W}^{(z)}$ \\
{${}$\hspace{4em} \bf else} keep previous $\mathcal{W}^{0}$ \\
{${}$\hspace{3em} \bf end if} \\
{${}$\hspace{2em} \bf reduce} temperature $\mathbf{T}_{old}$ based on (\ref{eq:16})-(\ref{eq:17}) \\
{${}$\hspace{2em} \bf end if} \\
{${}$\hspace{1em} \bf end} iteration $z$ \\
{\bf end} run epoch
\vspace{-0.1in} 
\end{algorithmic}
\line(1,0){245}
\vspace{-0.2in}  
\end{table}

\subsubsection{Acceptance of RBFN Weights.}
To avoid convergence to local minima, each run epoch (TTI) begins with an initial weight set $\mathcal{W}^0$, which may be updated over $I_t$ iterations to yield $\mathcal{W}^{(z)}$. The SAST-based management controller evaluates $\mathcal{A} = \mathcal{W}^{(z)}$, $\mathcal{B} = \mathcal{W}^0$, with $f = e$ and $f^* = e^*$. If $F(\mathcal{W}^{(z)}) \leq F(\mathcal{W}^0)$, then $\mathcal{W}^{(z)}$ is accepted as the initial set for the next run. Otherwise, if $F(\mathcal{W}^{(z)}) > F(\mathcal{W}^0)$, the non-better solution is accepted with probability $\mathbb{P}(\mathcal{W}^{(z)} | \mathcal{W}^0)$; it is retained only if this exceeds a random threshold $\mathbb{P}_{rand}$. If not, $\mathcal{W}^0$ is restored. The initial temperature $\mathbf{T}_0$ is computed over $L$ iterations as in (\ref{eq:16}), and gradually decreased using (\ref{eq:17}) at each TTI. Algorithm~II details the full RBFN-based CQI classification training process.

\vspace{-0.2in}
\subsection{Statistical Stage}
\label{sec:statistical}
At each TTI $t$, the classifier outputs the vector 
from \eqref{eq:24},
where $v_{M,k} \in [0,1]$ denotes the normalized number of users in class $k$, with $\sum\nolimits_k^K v_{M,k} \leq 1$ due to possible CQI delays or losses. Once parameters $\{M, K_M, \sigma, \eta\}$ are set and the RBF network is trained, classification can be offloaded to the mobile equipment. Reporting only the class index $k$ significantly reduces signaling overhead, while still enabling accurate CQI reconstruction at the base station when $K_M$ is sufficiently large. Additionally, dual reporting (class index and CQI vector) allows error correction by cross-referencing the two representations.

As shown in Fig.~\ref{figure_2}, the classification vector $\mathbf{v}_M$ is integrated into the controller state $\mathbf{s} \in \mathcal{S}$, alongside $\{\alpha_{t-1},\beta_{t-1}, \mathbf{\Gamma}[t], \Delta_t, \digamma_t, I_t\}$, to adjust the fairness parameters $[\alpha_t, \beta_t]$ at each TTI $t$ and ensure management feasibility. When the number of CQI classes $K_M$ increases, $\mathbf{v}_M$ may dominate the state, making feature extraction essential. The controller's decisions are guided by the sensitivity of $\alpha_t$ and $\beta_t$ to changes in $\mathbf{v}_M$ under NGMN fairness constraints. For instance, poor CQI distributions lead to $\alpha_t > \beta_t$ to favor fairness, whereas favorable distributions yield $\alpha_t < \beta_t$ to enhance throughput. Additionally, low dispersion in $\mathbf{v}_M$ permits reducing $\alpha_t$ to optimize capacity, while high dispersion requires increasing it to maintain fairness compliance.

To reduce the dimensionality of the classification vector $\mathbf{v}_M$ and retain its key characteristics, we extract a set of descriptive features. Let $h_k[t] \in \{0,1\}$ indicate if class $k$ is active at TTI $t$, such that $h_k[t]=1$ if $v_{M,k}>0$, and $h_k[t]=0$ otherwise. The total number of active classes is:
\vspace{-0.05in}
\begin{equation}
\label{eq:41a}
h_a = \sum\nolimits_{k=1}^{K_M} h_k.
\tag{41.a}
\vspace{-0.05in}
\end{equation}
The class dispersion is measured by:
\vspace{-0.05in}
\begin{equation}
\label{eq:41b}
\sigma_{a} = 1/h_a \cdot \sum\nolimits_{k=1}^{K_M} h_k \cdot (v_{M,k} - 1/h_a)^{2}.
\tag{41.b}
\vspace{-0.05in}
\end{equation}
To quantify the CQI quality over all vehicular users every TTI, we define a support set $\mathcal{Q} = \{\bar{\mathbf{v}}_k\}$, where each $\bar{\mathbf{v}}_k$ is a one-hot vector. The best match to the current $\mathbf{v}_M$ is found by minimizing the Euclidean distance:
\vspace{-0.05in}
\begin{equation}
\label{eq:41c}
d_{M,k} = ||\mathbf{v}_M - \bar{\mathbf{v}}_k ||, k=1,2,...,K_M,
\tag{41.c}
\vspace{-0.05in}
\end{equation}
and then, let $k_{\bar{\mathbf{v}}}$ be the index of the nearest support vector
\vspace{-0.05in}
\begin{equation}
\label{eq:41d}
k_{\bar{\mathbf{v}}} = argmin_{k^{\prime}} d(\mathbf{v}_M, \bar{\mathbf{v}}_{k^{\prime}}),
\tag{41.d}
\vspace{-0.05in}
\end{equation}
and $d_{M,k}$ the associated distance. 
\textcolor{blue}
Based on the proposed compression mechanism, the full vector $\mathbf{v}_M$ is replaced by the compact CQI feature vector
$[h_a,\sigma_a,d_{M,k},k_{\bar{\mathbf{v}}}]$.
Accordingly, the controller state in (\ref{eq:original_state}) becomes:
\vspace{-0.05in}
\begin{equation}
\label{eq:42}
\mathbf{s}[t]=[\alpha_{t-1},\beta_{t-1},
h_a,\sigma_a,d_{M,k},k_{\bar{\mathbf{v}}},
\mu(\mathbf{\Gamma}),\sigma(\mathbf{\Gamma}),
\Delta_t,\digamma_t,\bar{I}_t],
\tag{42}
\vspace{-0.05in}
\end{equation}
where $\mu(\mathbf{\Gamma})$ and $\sigma(\mathbf{\Gamma})$ denote the mean and standard deviation of user throughput, respectively, and $\bar{I}_t=I_t/I_{\max}$ is the normalized number of active users, where $I_{\max}$ denotes the maximum number of users that can be accommodated within the available bandwidth represented by $\mathcal{J}$. Equation~(\ref{eq:42}) defines the fixed-dimensional controller state for CQI compression configurations $(M,K)$. For the \textit{No Compression} baseline configuration, the CQI vector $[h_a,\sigma_a,d_{M,k},k_{\bar{\mathbf{v}}}]$ is omitted, while the remaining state variables are retained. Thus, in both cases, the RL agents adapt their scheduling policies to varying user populations through $\bar{I}_t$, without requiring variable-dimensional inputs.

The compressed CQI descriptors are not assumed to be statistically sufficient for the NGMN fairness constraint. Instead, they provide a low-dimensional state abstraction capturing occupancy, dispersion, and dominant features of the classified CQI distribution relevant to RL-based decisions. Fairness evaluation remains based on the user-throughput distribution and NGMN feasibility conditions, while the compressed CQI representation supplies contextual information for adapting fairness-control parameters under varying mobility and load. Thus, the compression objective is not lossless CQI reconstruction, but the construction of a compact, fixed-dimensional, and informative control-state representation for RL-based vehicular network management. The parameters of the compression pipeline serve distinct roles. $M$ and $K$ define the dimensionality and granularity of the compressed CQI representation used by the RL controller, while $\sigma$ and $\eta$ related to RBFN training, and $\omega$ and $R_T$ to SAST-based optimization. These latter parameters influence convergence and classification accuracy but do not directly affect scheduling aggressiveness. The throughput--fairness trade-off is controlled by the scheduler parameters $[\alpha_t,\beta_t]$.

\begin{table}[t]
\centering
\footnotesize 
\caption{\small \textbf{Parameter Settings}}
\label{table:table_1}
\begin{tabular}{p{4cm} p{3.5cm}} \hline
\textbf{Parameter} & \textbf{Value/Description} \\ \hline

\multicolumn{2}{c}{\textbf{Network Parameters}} \\ \hline
Downlink Transmission Power & 43dBm \\
Multipath Fading Model & Jakes Model with 12 Paths \\
Path Loss Model & Macro Cell Urban Area \\
Penetration Loss & 10 dB \\
Shadowing Loss Mean/Deviation & $\mu = 0$, $\sigma = 8dB$ \\
Noise Figure & 2.5 \\
Noise Spectral Density & -174dBm \\
Number of Cells / Reuse Factor & 19/3  \\
RB Bandwidth & 180KHz  \\
CQI Reporting Mode & Periodic, every 1ms \\
PUCCH Channel & Errorless \\
Number of CQI Feedbacks per TTI & 1000 \\
User Speed & 120 km/h \\
User Mobility Model & Random Direction \\
Bandwidth & 20MHz \\
Target BLER & 10\% \\ 
CQI Preprocessing Mode & $M \in \{3, 4, 5\}$ \\
Collection Size $(U_M)$ & $U_3=33596$, $U_4=144179$, \\& $U_5=206473$ \\ \hline

\multicolumn{2}{c}{\textbf{SAST-based Clustering Parameters}} \\ \hline
Max Iterations per Run & 10 \\
Total Number of Iterations $(Z_C)$ & 1000 \\
Minimum Distortion $(\bar{d}_{min})$ & 0.1 \\
Acceptance Probability ($\mathcal{K}$) ($\mathbb{P}_0$) & 0.5 \\
Acceptance Probability (K-Means) & 0.5 \\
Temperature Running Length $(L)$ & 10 \\
Temp. Reduction Factor $(R_T)$ & 0.95 \\
Tunneling Parameter $(\omega)$ & 0.02 \\
Number of Centers $(K)$ & $K \in \{64, 128, 256, 512 \}$ \\ \hline

\multicolumn{2}{c}{\textbf{SAST-based RBFN Training Parameters}} \\ \hline
Max Iterations per Run & 1000 \\
Total Number of Iterations $(Z_C)$ & $10^7$ \\
Acceptance Probability ($\mathcal{V}$) ($\mathbb{P}_0$) & 0.8 \\
Acceptance Probability (Weights) & 0.8 \\
Temperature Running Length$(L)$& 50000 \\
Temp. Reduction Factor$(R_T)$ & 0.99 \\
Tunneling Parameter$(\omega)$ & 0.1 \\
RBFNN Output Layer Function & Tangent Hyperbolic \\
Gaussian Parameter $(\sigma)$ & see TABLE \ref{table:table_3} \\
Learning Rate $(\eta)$ & see TABLE \ref{table:table_3} \\ \hline

\multicolumn{2}{c}{\textbf{Intelligent Management Controller Parameters}} \\ \hline
Frame Structure & FDD  \\
CQI Reporting Mode & Full-band, periodic per TTI \\
PUCCH Model & Errorless \\
Scheduler Type & GPF-SP / GPF-DP \\
Throughput Forgetting Factor & $\lambda_{\mathrm{f}} = 0.99$ \\
Traffic Type & Best Effort \\
RLC ARQ & No Retransmissions \\
Confidence Interval ($\Delta^{\mathrm{NGMN}}$) & 0.05 \\
Controller Timescale &  1 TTI \\
Number of Users $(I_t)$ & Variable: 15-120 \\
RL Algorithms & Q-L, DoubleQ-L, SARSA, \\& QV, QV2, QVMAX, \\& QVMAX2, ACLA, CACLA1, \\ &CACLA2 \\
Training/Testing Time & 3000s/200s (in average) \\
Neural Network Type & Feed-Forward \\
No. of hidden layers / nodes & 1/60 \\ \hline
\end{tabular}
\vspace{-0.2in}
\end{table}

\begin{table*}[t]
\centering
\footnotesize
\caption{Offline Clustering Performance (Average Distortion / CPU Time): KN in Absolute Values (CPU Time in Seconds); Other Methods as Relative Percentage Changes to KN.}
\label{table:table_2}
\begin{tabular}{ccccccc}
\toprule
\multirow{2}{*}{\textbf{Method}} & \multicolumn{2}{c}{\textbf{M = 3}} & \multicolumn{2}{c}{\textbf{M = 4}} & \multicolumn{2}{c}{\textbf{M = 5}} \\
\cmidrule(lr){2-3} \cmidrule(lr){4-5} \cmidrule(lr){6-7}
& \textbf{K = 64} & \textbf{K = 128} & \textbf{K = 64} & \textbf{K = 128} & \textbf{K = 64} & \textbf{K = 128} \\
\midrule
\textbf{KN} &  0.0193 / 16.09  &  0.0094 / 25.96  &  0.0111 / 56.9  &  0.0065	/ 90.25  &  0.0084 / 137.19  & 0.0058 / 208.18  \\
\textbf{RS} &  28.1\% / -1.2\%  & 30.4\% / 0.4\%  &  31.2\%	/ 4.5\%  & 32.8\% / 2.7\% & 26.4\% / -1.1\% & 25.2\% / 0.9\%  \\
\textbf{RSKN} & -1.7\% / -1.6\%  & 1.6\% / -1.4\%  &  3.4\% / -3.8\%  & 2.5\% / -3.4\%  &  0.9\% / -3.1\%  &  0.2\%	/ -2.3\%  \\
\textbf{SA} & -7.1\% / -3.3\%  & -8.1\% / -0.9\% & -2.4\%	/ -4.1\%  & -0.8\% / -3.6\% & -1.5\% / -5.7\% & -2.5\% /-3.5\%  \\
\textbf{SAST} & -7.3\% / -3.4\%  & -8.2\% / -0.6\%  &  -2.6\%	/ -6.8\%  & -1.7\% / -1.4\%  & -1.7\% / -6.8\%  &  -3.3\%	/ -4.9\%  \\
\bottomrule

\multirow{2}{*}{\textbf{Method}} & \multicolumn{2}{c}{\textbf{M = 3}} & \multicolumn{2}{c}{\textbf{M = 4}} & \multicolumn{2}{c}{\textbf{M = 5}} \\
\cmidrule(lr){2-3} \cmidrule(lr){4-5} \cmidrule(lr){6-7}
& \textbf{K = 256} & \textbf{K = 512} & \textbf{K = 256} & \textbf{K = 512} & \textbf{K = 256} & \textbf{K = 512} \\
\midrule
\textbf{KN} &  0.0045 /	43.44  & 0.0022 / 75.01  &  0.004 / 143.03  & 0.0027 / 223.8  &  0.0041	/ 320.52  &  0.0029	/ 504.21  \\
\textbf{RS} & 34.3\% / 1.8\%  & 54.3\% / 2.5\%  &  29.4\%	/ 1.4\% & 32.5\% / 0.3\% & 24.4\% /-0.9\% & 27.5\% / -2.7\%  \\
\textbf{RSKN} & 3.2\% / -0.4\% & -0.5\% / 0.8\% & -0.8\% / -3.5\%  & -1.2\% / -1.3\%  & 0.1\% / -1.1\% & -0.8\% / -1.8\%  \\
\textbf{SA} & -7.4\% / -0.4\%  & -5.7\% / 1.5\% & -4.1\% / -2.9\%  & -3.5\% / -1.9\% & -3.0\% / -2.3\% & -2.8\%	/ -1.6\%  \\
\textbf{SAST} & -9.4\% /-0.2\%  & -5.9\% / 0.7\%  &  -4.3\%	/ -2.2\%  & -3.9\% / -1.9\%  & -3.2\% / -2.3\% & -3.3\% / -1.9\%  \\
\bottomrule
\end{tabular}
\vspace{-0.1in}
\end{table*}

\vspace{-0.15in}
\section{Simulation Results}
The evaluation is performed using an enhanced version of LTE-Sim \cite{piro_ltesim_2011}, an open-source event-driven LTE network simulator that provides detailed support for mobility management, CQI reporting, radio-resource allocation, traffic generation, and scheduling procedures. LTE-Sim was selected due to its modular architecture and source-code accessibility, which enabled the integration of the proposed CQI compression, machine-learning, and reinforcement-learning components required by the SCAR framework. The enhanced LTE-Sim network simulator \cite{comsa_thesis_2014} supports the proposed CQI compression framework under two main settings: \textit{Online} (real-time processing) and \textit{Offline} (algorithm optimization). These settings are divided into four functional modes: (1) \textit{CQI Collection \& Pre-processing (Online)}: Gathers and processes CQI data for unique, representative samples (Section~\ref{sec:preprocessing}). (2) \textit{Clustering Analysis (Offline)}: Applies the SAST-based metaheuristic for offline clustering and benchmarking (Section~\ref{sec:clustering}). (3) \textit{RBFN Training (Online)}: Trains and updates the RBF network using dynamic CQI data (Section~\ref{sec:rbfn}). (4) \textit{Management \& Evaluation (Online)}: Employs compressed state representations in the RL-based management controller to adapt the throughput--fairness trade-off according to the NGMN-defined feasibility conditions (Sections~\ref{sec:scheduling},~\ref{sec:statistical}). All simulations were conducted on ten physical machines, each equipped with a 64-bit Intel Core i7-2600 quad-core processor (3.40~GHz), 8~GB RAM, and a 120~GB Western Digital HDD.

\vspace{-0.15in}
\subsection{Data Collection}
We simulate a 20~MHz cellular system ($J=100$) with 19 cells and 1000 users moving at 120~km/h under Jakes fading, as detailed in Table~\ref{table:table_1}. CQI reports are periodically and reliably received. To reduce redundancy and complexity, we apply a pre-processing step selecting the top $M$ CQI features ($M=\{3,4,5\}$), followed by a uniqueness check. Smaller $M$ values yield compact state spaces, eliminating the need for clustering or classification. Data collection runs in parallel for each $M$, with a stopping criterion defined using:
\begin{equation}
\label{eq:43}
u_{M,t} = (1 - \nu) u_{M,t-1} + \nu \cdot \left[ I_t - \frac{U_{M,t} - U_{M,t-1}}{I_t} \right],
\tag{43}
\vspace{-0.05in}
\end{equation}
where $\nu = 0.005$. Collection stops when $u_{M,t} > 0.99$, signaling saturation. The resulting dataset sizes are: $U_3=33596$, $U_4=144179$, and $U_5=206473$, used in offline clustering and online RBFN training.

\vspace{-0.15in}
\subsection{Performance Analysis of Proposed SAST-based Clustering}
Based on the datasets $\mathcal{U}_3$, $\mathcal{U}_4$, and $\mathcal{U}_5$, the proposed SAST-based clustering method is applied to compute the corresponding data center sets $\mathcal{K}_3$, $\mathcal{K}_4$, and $\mathcal{K}_5$. Its performance is compared against several clustering baselines: KN-tree-based K-means (KN), random center selection (RS), a hybrid RS-KN approach (RSKN), and simulated annealing (SA). KN uses neighborhood-based centroid updates; RS selects centers randomly at each iteration; RSKN performs a single RS step per run before switching to KN updates; and SA minimizes the distortion function $F(\mathcal{X})$ via probabilistic transitions, similar to SAST but without stochastic tunneling.

In offline mode, each clustering algorithm runs for a maximum of $Z_C = 1000$ iterations, organized into 100 runs of 10 iterations each. For SAST and SA, the initial acceptance probability is $\mathbb{P}_0 = 0.5$, governing both RS-to-KN transitions and acceptance of non-improving solutions. A cooling rate of $R_T = 0.95$ gradually reduces randomness, favoring KN-based refinement in later stages. SAST introduces an early termination condition: if the distortion improvement within a run falls below a threshold ($d^{(z)} - d^0 < \bar{d}_{min} = 0.1$), the run ends early, enabling more frequent resets and efficient exploration. The tunneling factor $\omega = 0.02$ balances exploration and exploitation: larger values mimic RS behavior by over-accepting non-improving centers, while smaller values risk early convergence to KN-like local minima. This setting ensures convergence toward global optima within ~50 iterations. All parameter values are listed in Table \ref{table:table_1}.

The clustering performance is evaluated for center sizes $K \in \{64, 128, 256, 512\}$ across datasets $\mathcal{U}_3$, $\mathcal{U}_4$, and $\mathcal{U}_5$, assessing how different $(M, K)$ configurations affect algorithm behavior. The analysis focuses on distortion and CPU time to compare the clustering quality and computational runtime of each method, particularly under varying SAST parameter settings. Table~\ref{table:table_2} reports the average clustering distortion and offline CPU execution time for KN as absolute values (CPU time: 16.09--504.21s), while RS, RSKN, SA, and SAST are expressed as relative percentage changes in both metrics with respect to KN for each $(M,K)$ configuration. Across all configurations, the KN-tree implementation exhibits competitive computational runtime, with CPU times similar to the RS method despite its centroid refinement step. As expected, increasing $M$ or $K$ raises the complexity across all algorithms. Still, the proposed SAST method strikes the best balance between computational cost and clustering distortion. In terms of distortion, RS consistently performs worse due to its random center selection strategy, which lacks refinement. Hybrid methods improve upon KN, confirming its tendency to converge to local minima. RSKN, while simpler, offers moderate gains but lacks flexibility due to its fixed switching logic. In contrast, the SAST algorithm, with its dynamic tunneling mechanism, achieves the lowest distortion across all configurations. Notably, SAST's advantage grows with higher $K$ for a fixed $M$, but diminishes as $M$ increases. This trend indicates that low-dimensional pre-processing (e.g., $M=3$) provides an effective configuration for CQI clustering, offering low distortion without the additional computational cost associated with larger $M$.

\begin{figure*}[t]
\includegraphics[width=18cm]{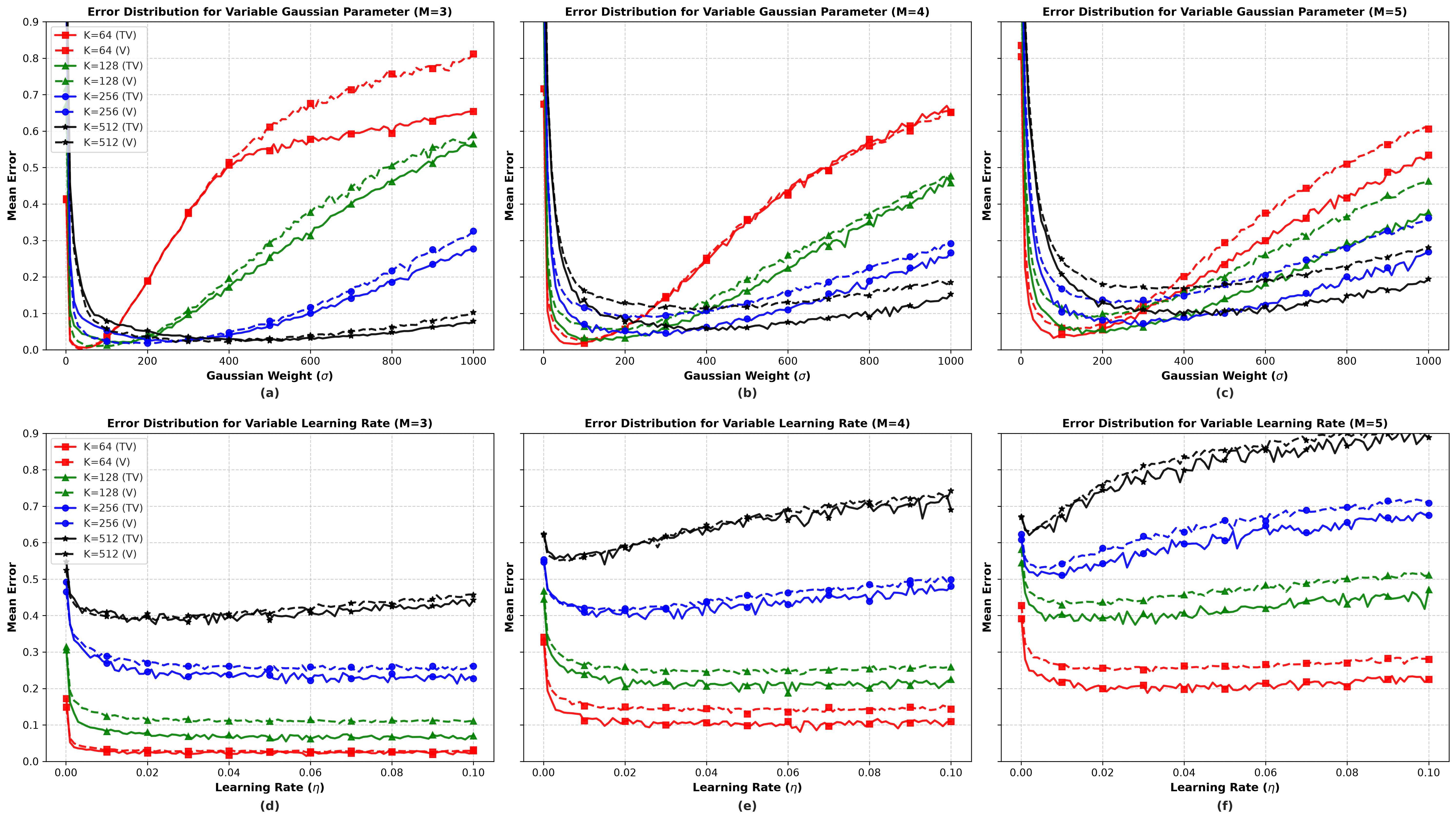}
\centering
\caption{The Error Distribution for the Gaussian Parameter ($\sigma$) when processing top a) $M=3$, b) $M=4$, c) $M=5$ CQI features. The Error Distribution for the Learning Rate ($\eta$) when processing top d) $M=3$, e) $M=4$, f) $M=5$ CQI features.}
\label{fig:figure_4}
\vspace{-0.2in}
\end{figure*}

\begin{table}[t]
    \centering
    \scriptsize
    \setlength{\tabcolsep}{3pt}
    \caption{\small Learning Rate, Gaussian Parameter, and RBFN Training CPU Time for Different $K$ and $M$ Values}
    \label{table:table_3}
    \begin{tabular}{c c c c c c c c c c}
        \toprule
        \textbf{K} &
        \multicolumn{3}{c}{\textbf{M=3}} &
        \multicolumn{3}{c}{\textbf{M=4}} &
        \multicolumn{3}{c}{\textbf{M=5}} \\
        \cmidrule(lr){2-4}
        \cmidrule(lr){5-7}
        \cmidrule(lr){8-10}
        & $\eta$ & $\sigma$ &
        CPU (s) &
        $\eta$ & $\sigma$ &
        CPU (s) &
        $\eta$ & $\sigma$ &
        CPU (s) \\
        \midrule
        64 &
        0.089 & 50 & 40.33 &
        0.05 & 90 & 43.50 &
        0.032 & 120 & 44.78 \\
        128 &
        0.075 & 130 & 80.93 &
        0.063 & 180 & 86.46 &
        0.01 & 180 & 87.94 \\
        256 &
        0.072 & 270 & 160.05 &
        0.021 & 210 & 168.41 &
        0.007 & 230 & 162.38 \\
        512 &
        0.022 & 440 & 342.06 &
        0.006 & 370 & 345.10 &
        0.002 & 310 & 336.42 \\
        \bottomrule
    \end{tabular}
    \vspace{-0.2in}
\end{table}

\vspace{-0.15in}
\subsection{Performance of Proposed SAST-based RBFN Training}
The sets of centers $\mathcal{K}_3$, $\mathcal{K}_4$, and $\mathcal{K}_5$, computed via SAST for $K \in \{64, 128, 256, 512\}$, are used as activation centers in separate RBF networks. Each RBFN is trained online for its corresponding configuration, enabling dynamic integration of historical CQI data ($\mathcal{U}_3$, $\mathcal{U}_4$, $\mathcal{U}_5$) with newly generated reports from the network simulator. This training strategy improves the model's generalization and supports robust classification of pre-processed CQI reports into quality patterns across diverse operational conditions.

To train the RBF networks, we simulate a vehicular network with $I_t = 1000$ users reporting CQI values at each TTI while moving at 120 km/h, using the configuration parameters from Table \ref{table:table_1}. The 1000 CQI reports received per TTI are pre-processed for $M \in \{3,4,5\}$ to form the training set $\mathcal{T}$ as in (\ref{eq:35}). To avoid overfitting, particularly when users experience prolonged similar channel conditions, the SAST metaheuristic is applied during training to escape local minima and improve weight updates. A total of $Z_R = 10^7$ iterations are performed, structured into 10000 runs (epochs) with 1000 iterations each (aligned with the number of CQI samples per TTI). Within each run, SAST adaptively switches between $\mathcal{T}$ and
$\mathcal{V}$ and decides whether to retain the current weight configuration for initialization in the next run, effectively balancing exploration and exploitation during training.

\begin{table*}[t]
\centering
\footnotesize
\caption{Performance Comparison of RL-based management policies (Mean Percentage of TTIs when the CDF is Located in the Unfair ($\mathcal{UF}$) / feasible ($\mathcal{FA}$) / overfair ($\mathcal{OF}$) Areas Together with the Confidence Intervals in Terms of Standard Deviation} 
\label{table:table_4}
\begin{tabular}{ccccccc}
\toprule
\multirow{2}{*}{\textbf{Method}} & \multicolumn{2}{c}{\textbf{M = 3}} & \multicolumn{2}{c}{\textbf{M = 4}} & \multicolumn{2}{c}{\textbf{M = 5}} \\
\cmidrule(lr){2-3} \cmidrule(lr){4-5} \cmidrule(lr){6-7}
& \textbf{K = 64} & \textbf{K = 128} & \textbf{K = 64} & \textbf{K = 128} & \textbf{K = 64} & \textbf{K = 128} \\
\midrule
\textbf{Q} &  11.03 / 86.66 / 2.31  & 5.53 / 89.59 / 4.88 & 2.84 / 76.44 / 20.73 & 8.2 / 65.69 / 26.11 & 6.25 / 79.86 / 13.89 & 11.36 / 85.8 / 2.84  \\
(STD) &  0.03 / 0.63 / 0.68 & 0.07 / 0.21 / 0.22 & 0.16 / 0.81 / 0.78 & 0.05 / 0.16 / 0.12 & 0.08 / 0.93 / 0.91 & 0.12 / 0.12 / 0.07 \\
\textbf{D-Q} &  12.04 / 56.96 / 31.01  & 3.74 / 80.84 / 15.42 & 9.51 / 54.82 / 35.67 & 5.69 / 67.83 / 26.48 & 9.82 / 86.94 / 3.24 & 7.34 / 88.74 / 3.93 \\
(STD) &  1.19 / 0.52 / 0.76 & 0.09 / 0.86 / 0.84 & 1.06 / 1.19 / 1.15 & 0.18 / 0.4 / 0.49 & 0.07 / 0.15 / 0.09 & 0.5 / 0.53 / 0.16 \\
\textbf{SARSA} &  1.94 / 87.96 / 10.11  & 2.66 / 87.16 / 10.18 & 2.47 / 90.84 / 6.69 & 2.02 / 90.74 / 7.24 & 2.54 / 92.78 / 4.69 & 5.85 / 90.48 / 3.67 \\
(STD) & 0.58 / 0.86 / 0.89 & 1.14 / 1.00 / 0.59 & 0.13 / 0.51 / 0.51 & 0.08 / 0.42 / 0.4 & 0.09 / 0.26 / 0.31 & 0.81 / 0.77 / 0.12 \\
\textbf{QV} &  3.43 / 90.53 / 6.04  & 1.78 / 94.57 / 3.66 & 5.96 / 91.98 / 2.06 & 6.62 / 88.74 / 4.65 & 6.07 / 92.49 / 1.45 & 1.95 / 93.25 / 4.8 \\
(STD) & 0.16 / 1.48 / 1.48 & 0.16 / 4.01 / 4.06 & 0.32 / 0.3 / 0.12 & 0.14 / 1.05 / 1.05 & 0.36 / 0.37 / 0.14 & 0.06 / 0.18 / 0.17 \\
\textbf{QV2} &  8.71 / 90.24 / 1.06  & 3.49 / 87.24 / 9.27 & 7.95 / 87.54 / 4.51 & 4.02 / 91.01 / 4.97 & 3.16 / 93.72 / 3.12 & 5.53 / 88.16 / 6.31 \\
(STD) & 0.49 / 0.49 / 0.05 & 0.08 / 0.2 / 0.16 & 0.52 / 0.61 / 0.13 & 0.09 / 0.61 / 0.62 & 0.28 / 0.4 / 0.34 & 0.45 / 0.48 / 0.27 \\
\textbf{QVMax} &  9.91 / 88.04 / 2.06  & 9.83 / 87.66 / 2.51 & 10.99 / 87.22 / 1.79 & 2.61 / 86.87 / 10.52 & 4.41 / 90.99 / 4.61 & 3.65 / 88.17 / 8.18 \\
(STD) &  0.07 / 0.09 / 0.06 & 0.79 / 0.67 / 0.32 & 0.05 / 0.14 / 0.11 & 0.07 / 0.22 / 0.17 & 0.07 / 0.17 / 0.11 & 0.05 / 0.56 / 0.55 \\
\textbf{QVMax2} &  3.64 / 93.79 / 2.56  & 7.44 / 91.68 / 0.89 & 8.37 / 89.44 / 2.19 & 4.38 / 93.12 / 2.49 & 1.79 / 94.22 / 1.43 & 3.62 / 92.99 / 3.38 \\
(STD) &  0.08 / 0.12 / 0.07 & 0.39 / 0.41 / 0.06 & 0.17 / 0.18 / 0.06 & 0.19 / 0.2 / 0.05 & 0.05 / 0.14 / 0.16 & 0.08 / 0.12 / 0.09 \\
\textbf{ACLA} &  4.61 / 90.88 / 4.51  & 3.36 / 92.69 / 3.95 & 3.33 / 94.46 / 2.22 & 4.76 / 92.74 / 2.5 & 4.98 / 93.58 / 4.59 & 3.36 / 93.32 / 3.32 \\
(STD) &  0.07 / 0.16 / 0.09 & 0.04 / 0.14 / 0.11 & 0.14 / 0.13 / 0.08 & 0.11 / 0.14 / 0.05 & 0.25 / 0.25 / 0.14 & 0.07 / 0.19 / 0.15 \\
\textbf{CACLA1} &  2.74 / 96.64 / 0.62  &  3.94 / 94.87 / 1.19 & 1.77 / 96.57 / 1.66 & 1.73 / 94.54 / 3.73 & 1.77 / 93.65 / 3.99 &  1.88 / 94.13 / 3.99 \\
(STD) & 0.06 / 0.07 / 0.04 & 0.24 / 0.19 / 0.27 & 0.05 / 0.11 / 0.08 & 0.04 / 0.19 / 0.19 & 0.03 / 0.48 / 0.48 & 0.04 / 0.25 / 0.26 \\
\textbf{CACLA2} &  2.11 / 96.78 / 1.11  & 2.12 / 96.16 / 1.73 & 1.69 / 95.23 / 3.08 & 2.84 / 96.52 / 0.64 & 1.69 / 94.64 / 3.67 & 1.74 / 96.45 / 1.81 \\
(STD) & 0.06 / 0.14 / 0.16 & 0.06 / 0.35 / 0.31 & 0.05 / 0.27 / 0.25 & 0.08 / 0.09 / 0.04 & 0.03 / 0.37 / 0.38 & 0.05 / 0.31 / 0.34 \\
\bottomrule
\multirow{2}{*}{\textbf{Method}}  & \multicolumn{2}{c}{\textbf{M = 3}} & \multicolumn{2}{c}{\textbf{M = 4}} & \multicolumn{2}{c}{\textbf{M = 5}} \\
\cmidrule(lr){2-3} \cmidrule(lr){4-5} \cmidrule(lr){6-7}
& \textbf{K = 256} & \textbf{K = 512} & \textbf{K = 256} & \textbf{K = 512} & \textbf{K = 256} & \textbf{K = 512} \\
\midrule
\textbf{Q} & 5.91 / 69.48 / 24.61 & 25.75 / 72.22 / 2.03 & 1.64 / 40.02 / 58.34 & 12.69 / 77.9 / 9.41 & 10.42 / 84.4 / 5.17 & 2.79 / 86.25 / 10.96 \\
(STD) & 0.13 / 0.95 / 0.95 & 0.19 / 0.19 / 0.07 & 0.05 / 0.29 / 0.3 & 0.72 / 0.63 / 0.36 & 0.44 / 0.48 / 0.13 & 0.05 / 0.09 / 0.06 \\
\textbf{D-Q} & 10.87 / 86.78 / 2.36 & 2.84 / 44.2 / 52.96 & 6.78 / 88.93 / 4.29 & 7.44 / 74.67 / 17.89 & 1.67 / 44.97 / 53.36 & 4.23 / 90.15 / 5.62 \\
(STD) & 0.03 / 0.19 / 0.19 & 0.04 / 0.42 / 0.43 & 0.11 / 0.17 / 0.09 & 0.05 / 0.21 / 0.21 & 0.06 / 0.94 / 0.96 & 0.06 / 0.3 / 0.26 \\
\textbf{SARSA} & 1.7 / 84.77 / 13.53 & 35.25 / 58.13 / 6.63 & 1.72 / 87.37 / 10.91  & 1.68 / 83.74 / 14.58 & 2.74 / 87.69 / 9.56 & 8.07 / 89.83 / 2.09 \\
(STD) &  0.04 / 0.55 / 0.57 & 2.71 / 2.57 / 0.58 & 0.06 / 0.65 / 0.67 & 0.04 / 0.86 / 0.88 & 0.15 / 0.24 / 0.19 & 0.72 / 0.7 / 0.06 \\
\textbf{QV} & 4.01 / 94.02 / 1.97 & 8.49 / 91.13 / 0.39 & 2.19 / 92.51 / 5.3 & 5.25 / 89.72 / 5.02 & 5.81 / 92.05 / 2.14 & 1.72 / 95.65 / 2.64 \\
(STD) & 0.19 / 0.23 / 0.07 & 0.21 / 0.22 / 0.07 & 0.04 / 0.16 / 0.15 & 0.13 / 0.24 / 0.17 & 0.15 / 0.2 / 0.09 & 0.03 / 0.29 / 0.31 \\
\textbf{QV2} & 8.93 / 88.59 / 2.48 & 3.29 / 91.39 / 5.31 & 11.04 / 85.07 / 3.89 & 3.12 / 93.21 / 3.67 & 5.04 / 93.03 / 1.93 & 2.11 / 95.09 / 2.8 \\
(STD) & 0.16 / 0.2 / 0.12 & 0.06 / 0.24 / 0.21 & 0.1 / 0.33 / 0.29 & 0.07 / 0.09 / 0.07 & 0.11 / 0.12 / 0.11 & 0.06 / 0.09 / 0.05 \\
\textbf{QVMax} & 1.7 / 64.59 / 33.71 & 2.49 / 86.17 / 11.34 & 2.16 / 90.42 / 7.41 & 5.8 / 89.97 / 4.23 & 6.58 / 89.56 / 3.86 & 2.0 / 88.49 / 9.51 \\
(STD) & 0.03 / 0.09 / 0.07 & 0.07 / 0.62 / 0.65 & 0.2 / 0.35 / 0.24 & 0.09 / 0.16 / 0.11 & 0.39 / 0.37 / 0.09 & 0.13 / 0.35 / 0.36 \\
\textbf{QVMax2} & 2.77 / 93.39 / 3.85 & 2.46 / 93.42 / 4.12 & 2.71 / 92.26 / 5.03 &  2.08 / 95.51 / 2.41 & 6.33 / 88.3 / 5.37 & 3.98 / 94.55 / 1.47 \\
(STD) &  0.15 / 0.16 / 0.09 & 0.1 / 0.38 / 0.36 & 0.29 / 0.34 / 0.23 & 0.07 / 0.09 / 0.06 & 0.85 / 0.92 / 0.12 & 0.12 / 0.14 / 0.04 \\
\textbf{ACLA} & 4.82 / 93.13 / 2.06 & 4.08 / 94.19 / 1.73 & 3.89 / 92.89 / 3.22 & 4.18 / 92.06 / 3.76 & 3.09 / 93.79 / 3.12 & 7.3 / 91.48 / 1.22 \\
(STD) & 0.06 / 0.1 / 0.04 & 0.08 / 0.15 / 0.11 & 0.09 / 0.15 / 0.09 & 0.09 / 0.11 / 0.06 & 0.11 / 0.76 / 0.83 & 0.11 / 0.13 / 0.04 \\
\textbf{CACLA1} & 1.87 / 96.54 / 1.59 & 5.43 / 93.92 / 0.65 & 1.88 / 97.34 / 0.79 &  1.8 / 93.43 / 4.77 & 1.94 / 96.55 / 1.51 & 1.89 / 95.66 / 2.45 \\
(STD) & 0.04 / 0.25 / 0.25 & 0.6 / 0.55 / 0.13 & 0.2 / 0.25 / 0.13 & 0.04 / 0.34 / 0.34 & 0.05 / 0.09 / 0.1 & 0.21 / 0.22 / 0.13 \\
\textbf{CACLA2} & 2.06 / 94.89 / 3.04 & 1.83 / 96.48 / 1.69 & 1.89 / 97.54 / 0.58 & 1.88 / 97.76 / 0.37 & 1.98 / 96.3 / 1.72 & 1.75 / 97.42 / 0.83 \\
(STD) & 0.04 / 0.18 / 0.17 & 0.03 / 0.07 / 0.07 & 0.08 / 0.07 / 0.05 & 0.04 / 0.05 / 0.03 & 0.3 / 0.36 / 0.1 & 0.03 / 0.1 / 0.11 \\
\bottomrule
\end{tabular}
\vspace{-0.2in}
\end{table*}

\begin{table}[t]
\centering
\footnotesize
\caption{Average Performance of RL-Based Management Policies
With and Without CQI Compression ($\mathcal{UF}$ / $\mathcal{FA}$ / $\mathcal{OF}$, in \%). Compressed Results Are Averaged Across All 12 $(M,K)$ Configurations.}
\label{table:rl_compression_summary}
\begingroup
\begin{tabular}{lcc}
\toprule
\textbf{Method} &
\textbf{No Compression} &
\textbf{CQI Compression} \\
\midrule
\textbf{Q}
& 1.92 / 13.05 / 85.03
& 8.70 / 76.19 / 15.11 \\
(STD)
& 0.18 / 0.60 / 0.49
& 6.50 / 13.64 / 16.10 \\

\textbf{D-Q}
& 2.39 / 28.79 / 68.83
& 6.83 / 72.15 / 21.02 \\
(STD)
& 0.28 / 0.17 / 0.14
& 3.31 / 17.75 / 18.89 \\

\textbf{SARSA}
& 2.88 / 69.68 / 27.44
& 5.72 / 85.96 / 8.32 \\
(STD)
& 0.40 / 1.54 / 1.55
& 9.50 / 9.14 / 3.83 \\

\textbf{QV}
& 11.37 / 80.34 / 8.30
& 4.44 / 92.22 / 3.34 \\
(STD)
& 0.85 / 6.70 / 6.73
& 2.25 / 2.01 / 1.80 \\

\textbf{QV2}
& 60.76 / 38.91 / 0.33
& 5.53 / 90.36 / 4.11 \\
(STD)
& 0.67 / 0.68 / 0.02
& 2.90 / 3.08 / 2.20 \\

\textbf{QVMax}
& 26.28 / 73.17 / 0.55
& 5.18 / 86.51 / 8.31 \\
(STD)
& 2.16 / 2.18 / 0.11
& 3.41 / 7.06 / 8.67 \\

\textbf{QVMax2}
& 8.09 / 22.81 / 69.11
& 4.13 / 92.72 / 2.93 \\
(STD)
& 0.98 / 1.07 / 0.74
& 2.15 / 2.07 / 1.43 \\

\textbf{ACLA}
& 9.87 / 82.97 / 7.15
& 4.31 / 92.93 / 3.02 \\
(STD)
& 0.81 / 1.57 / 0.87
& 1.15 / 1.06 / 1.09 \\

\textbf{CACLA1}
& 11.13 / 70.79 / 18.08
& 2.39 / 95.32 / 2.24 \\
(STD)
& 0.54 / 1.35 / 0.94
& 1.15 / 1.38 / 1.49 \\

\textbf{CACLA2}
& 16.90 / 82.73 / 0.37
& 1.96 / 96.35 / 1.69 \\
(STD)
& 0.50 / 0.58 / 0.11
& 0.32 / 1.01 / 1.08 \\

\midrule
\textbf{Overall Average}
& \textbf{15.16 / 56.32 / 28.52}
& \textbf{4.92 / 88.07 / 7.01} \\
\textbf{(STD)}
& \textbf{17.65 / 27.31 / 33.02}
& \textbf{1.98 / 8.10 / 6.42} \\
\bottomrule
\end{tabular}
\endgroup
\vspace{-0.25in}
\end{table}

To facilitate stable RBFN training, the initial acceptance probability for both data switching and weight updates is set to $\mathbb{P}_0 = 0.99$, with a temperature decay rate of $R_T = 0.99$, enabling gradual cooling across the high iteration count. A tunneling factor $\omega = 0.1$ ensures broad exploration over 10 training runs (10,000 samples). Early training emphasizes exploration by favoring the dataset $\mathcal{T}$ and accepting suboptimal weights, promoting diversity in the weight space. As training advances, SAST shifts focus to validation data $\mathcal{V}$ and tightens weight acceptance, allowing only improved solutions to persist. This adaptive behavior guides the network toward convergence, with full training parameters listed in Table \ref{table:table_1}.

Before training the RBF networks, optimal values for the Gaussian parameter $(\sigma)$ and learning rate $(\eta)$ must be determined for each $(M,K)$ configuration. Figure~\ref{fig:figure_4} compares error distributions for the proposed SAST-based training (dashed lines) and conventional training on $\mathcal{V}$ alone (solid lines). As expected, SAST yields slightly higher mean errors due to its controlled exploration, which prevents premature convergence and enhances generalization by dynamically integrating both $\mathcal{T}$ and $\mathcal{V}$ samples. Figures~\ref{fig:figure_4}.(a)–(c) show error trends for varying $\sigma$ (with fixed $\eta=0.1$) across $M \in \{3,4,5\}$. Two key observations emerge: (i) increasing the number of centers $K$ reduces error due to improved model expressiveness; (ii) higher $M$ values also reduce error, reflecting the benefits of richer input representations. Notably, optimal $\sigma$ values shift upward as $M$ and $K$ increase, indicating a need for broader basis functions to ensure smooth generalization in more complex models. Figures~\ref{fig:figure_4}.(d)–(f) analyze error for varying $\eta$ (with fixed $\sigma=10$). Here, increasing $K$ leads to higher errors, likely due to over-complexity and limited training data per center. Likewise, increasing $M$ introduces greater input granularity but also raises sensitivity to preprocessing accuracy, slightly elevating error. To identify the best ($\sigma, \eta$) pair for each $(M,K)$, extensive validation is performed, and the optimal configurations are listed in Table~\ref{table:table_3}. This table also reports the accumulated RBFN training CPU time over 10.000 TTIs, considering 1.000 CQI reports per TTI. Training CPU time increases approximately linearly with $K$, whereas variations across $M$ are comparatively small. These settings are used in the next subsection to compress the CQI states in RL-based management controller.

\vspace{-0.15in}
\subsection{Performance Analysis of RL-Assisted Management Policies}
This study analyzes the performance of RL-assisted management policies operating under NGMN fairness constraints, using compressed CQI inputs generated by RBF networks. Different $(M, K)$ configurations are tested to evaluate how CQI compression affects the RL agents’ ability to guide the management controller toward the feasible region $\mathcal{FA}$, while minimizing time spent in unfair $\mathcal{UF}$ and overfair $\mathcal{OF}$ regions. The goal is to assess the long-term effectiveness of compressed-state RL policies versus their uncompressed counterparts and state-of-the-art approaches in terms of fairness performance. The simulation scenario features a fast-varying mobile environment with users traveling at 120 km/h. User activity dynamically ranges between 15 and 120 devices, randomly switching between online and idle states. The system employs an FDD structure with periodic full-band CQI reporting and a downlink-only, best-effort traffic model to isolate fairness behavior from traffic-dependent variations. The RL-based policies are based on the GPF rule, with fairness control handled via either simple parameterization (SP), adjusting a single fairness parameter $\alpha_t$, or double parameterization (DP), where both $\alpha_t$ and $\beta_t$ are jointly optimized. To train and evaluate the proposed system, a diverse set of RL algorithms is applied, including Q-Learning, SARSA, Double-Q, QV variants, ACLA, and CACLA \cite{comsa_thesis_2014}. CACLA2 supports continuous DP-based control, while all other RL approaches are used for continuous SP adaptation. These models are trained for $3\cdot10^{6}$ TTIs with various $(M, K)$ compression settings and are evaluated over ten $2\cdot10^{5}$ TTIs test runs under consistent conditions. Unless otherwise stated, the NGMN confidence interval is set to $\Delta^{\mathrm{NGMN}} = 0.05$. Preliminary experiments showed that smaller intervals produce excessively sparse reward signals due to the limited number of feasible states, while larger intervals generate overly broad feasible regions and reduce the sensitivity of the reward mechanism. The selected value therefore provides a practical balance between reward availability and sensitivity to the NGMN-defined fairness conditions. The results, averaged across trials, provide a robust comparison of how CQI compression affects the fairness-oriented performance of RL-based RRM policies.

\begin{figure*}[t]
\includegraphics[width=17cm]{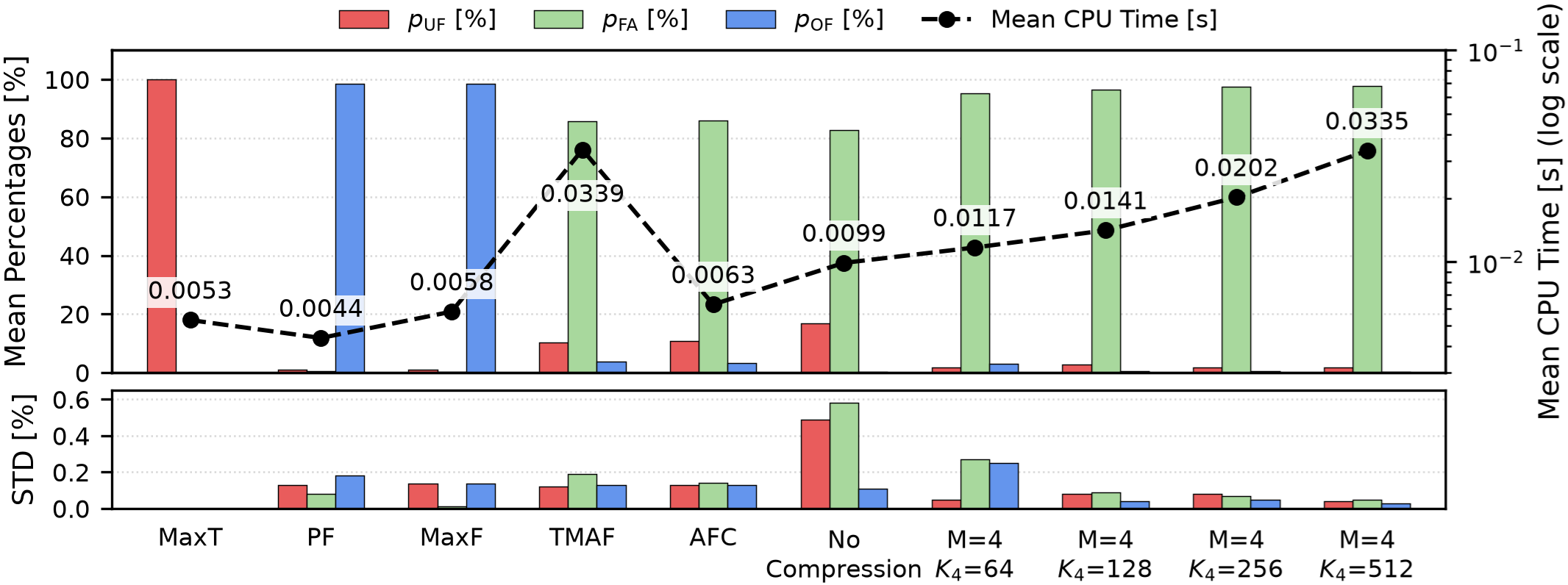}
\centering
\caption{Ablation Study on CQI Compression for CACLA2 and Performance Comparison With State-of-the-Art Resource Management Methods}
\vspace{-0.1in}
\label{fig:figure_5}
\vspace{-0.1in}
\end{figure*}

Table \ref{table:table_4} summarizes the performance of the evaluated RL-based management policies under different CQI-compression configurations $(M,K)$, reporting the mean percentage of TTIs spent in the unfair ($p_{uf}$), feasible ($p_{fa}$), and overfair ($p_{of}$) regions across 10 independent simulation runs, together with the corresponding standard deviations. Overall, the actor-critic methods consistently outperform value-based approaches in terms of fairness compliance. In particular, CACLA2 achieves the best performance and robustness, reaching a maximum feasible-region occupancy of $97.76\%$ for $(M=4,K=512)$ while maintaining very low unfair and overfair operation and exhibiting minimal variability across runs. In contrast, value-based methods show greater sensitivity to the compression configuration, often resulting in lower feasible-region occupancy. The results also indicate that increasing $K$ generally benefits actor-critic methods, whereas performance gains from increasing $M$ beyond 4 are marginal. CACLA2 with $(M=4,K_4=512)$ emerges as the most effective and statistically robust configuration among the evaluated approaches.

Table \ref{table:rl_compression_summary} compares the performance of ten RL-based scheduling policies with and without CQI compression, reporting the percentages of time spent in the unfair ($\mathcal{UF}$), feasible ($\mathcal{FA}$), and overfair ($\mathcal{OF}$) regions, together with their standard deviations. Compressed results are averaged across all 12 $(M,K)$ configurations. Overall, CQI compression increases the average feasible-region occupancy from 56.32\% to 88.07\%, while reducing unfair allocation time from 15.16\% to 4.92\% and overfair allocation time from 28.52\% to 7.01\%. Improvements are observed in feasible-region occupancy across all ten algorithms, although their magnitude varies. CACLA2 achieves the highest average feasible-region occupancy (96.35\%) and the lowest unfair allocation time (1.96\%) under compression. These findings indicate that compact CQI state representations generally improve fairness-oriented RL radio resource management across different learning algorithms.

Figure~\ref{fig:figure_5} compares CACLA2 ($K_4 \in \{64,128,256,512\}$, $M=4$) with the conventional scheduling rules MaxT, PF, and MaxF, as well as the state-of-the-art approaches, Maximizing Throughput with Adjustable Fairness (TMAF) \cite{tmaf_schwarz}, and Adaptive Fairness Control (AFC) \cite{afc_proebster}. The reported values correspond to the mean percentages of TTIs spent in the unfair ($p_{uf}$), feasible ($p_{fa}$), and overfair ($p_{of}$) regions, together with their standard deviations across 10 independent simulation runs of $2 \cdot 10^{5}$ TTIs each. MaxT operates almost exclusively in the unfair region ($p_{uf}>99.9\%$), prioritizing throughput at the expense of fairness, whereas MaxF and PF exhibit excessive overfairness ($p_{of}>98\%$). In contrast, TMAF and AFC operate within the feasible region for approximately 85\% of TTIs. CACLA2 without CQI compression achieves only 82.73\% feasible-region occupancy, performing worse than both TMAF and AFC. However, the proposed CQI compression mechanism substantially improves performance, with CACLA2 ($M=4$, $K_4=512$) increasing feasible-region occupancy to 97.76\% while reducing unfair and overfair operation to 1.88\% and 0.37\%, respectively. The compressed configurations also exhibit consistently low standard deviations, indicating stable behavior across independent runs.

The reported CPU times (Fig.~\ref{fig:figure_5}) represent the mean execution time per TTI of the complete scheduling procedure, obtained by averaging the accumulated computational time across TTIs and independent simulation runs. This includes scheduling and resource allocation for MaxT, PF, and MaxF; fairness-parameter determination followed by scheduling and resource allocation for TMAF and AFC; RL inference, scheduling, and resource allocation for uncompressed CACLA2; and additional CQI aggregation for compressed CACLA2. These measurements characterize computational overhead rather than compliance with the 1-ms TTI deadline. PF exhibits the lowest runtime (0.0044s), followed by MaxT, MaxF, and AFC, whereas TMAF incurs higher overhead (0.0339s). For CACLA2, computational overhead increases with the number of RBF centers ($K$). Nevertheless, CACLA2 ($M=4$, $K_4=512$) achieves a comparable runtime (0.0335s) to TMAF while substantially improving feasible-region occupancy (97.76\% versus 85.86\%).

\begin{figure*}[t]
\includegraphics[width=18.1cm]{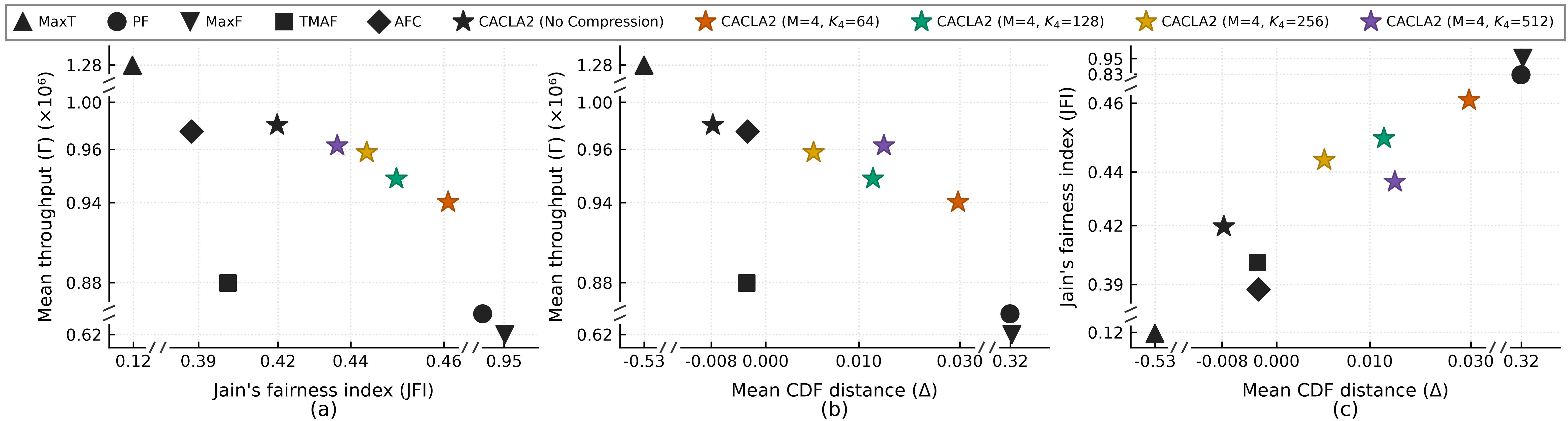}
\centering
\caption{Trade-offs Across Conventional Scheduling Algorithms and CACLA2 Compression Configurations: (a) Throughput ($\Gamma$) vs. JFI; (b) Throughput ($\Gamma$) vs. CDF distance ($\Delta$); (c) JFI vs. CDF distance ($\Delta$).}
\vspace{-0.1in}
\label{fig:figure_6}
\vspace{-0.15in}
\end{figure*}

Figure \ref{fig:figure_6} examines the trade-offs among mean user throughput ($\Gamma$), Jain's fairness index (JFI), and signed CDF distance ($\Delta$), with values averaged across 10 independent simulation runs. Figure \ref{fig:figure_6}(a) shows an inverse relationship between throughput and JFI. MaxT achieves the highest throughput but lowest fairness, whereas PF and MaxF exhibit the opposite behavior. TMAF, AFC, and CACLA2 provide intermediate operating points. For compressed CACLA2, increasing $K$ improves throughput at the expense of moderately reduced JFI, reflecting a controllable throughput-fairness trade-off. Figure \ref{fig:figure_6}(b) reveals that MaxT, TMAF, AFC, and uncompressed CACLA2 exhibit negative CDF distances, indicating unfair operation, whereas PF and MaxF exceed the NGMN tolerance threshold ($\Delta^{\mathrm{NGMN}}=0.05$), indicating overfairness. Only compressed CACLA2 configurations achieve small positive distances within this threshold while maintaining high throughput. Among them, $K_4=256$ achieves the smallest positive deviation, whereas $K_4=512$ provides the highest throughput. Figure \ref{fig:figure_6}(c) identifies an empirically favorable JFI range of approximately $0.43$--$0.47$ for compressed CACLA2, corresponding to NGMN-compliant mean CDF distances. Although $K_4=64$ achieves the highest JFI among these configurations, it also exhibits over 3\% overfair-region occupancy. Increasing $K$ generally reduces both JFI and overfair occupancy, reaching only 0.37\% for $K_4=512$. Thus, moderately reducing JFI can improve NGMN fairness regulation by limiting excessive resource equalization while maintaining operation within the feasible region.

The trade-offs presented in Fig. \ref{fig:figure_6} are consistent with the feasible-region occupancy results in Fig. \ref{fig:figure_5}. MaxT operates predominantly in the unfair region, while PF and MaxF remain largely overfair. TMAF and AFC achieve approximately 85\% feasible occupancy, compared with 82.73\% for uncompressed CACLA2. In contrast, CACLA2 ($M=4$, $K_4=512$) achieves 97.76\% feasible occupancy, with only an approximately 1.8\% throughput reduction relative to its uncompressed counterpart. Despite the additional CQI aggregation and RL inference, its computational overhead (0.0335s) remains comparable to TMAF (0.0339s). Overall, CACLA2 ($M=4$, $K_4=512$) provides the most favorable trade-off among the evaluated configurations, maintaining high throughput while satisfying the prescribed fairness constraints for the largest proportion of TTIs at competitive computational cost.

\vspace{-0.1in}
\section{Conclusions and Future Research Directions}
This paper presented SCAR (State-Space Compression for AI-Based Radio Resource Management), a framework for compressing high-dimensional CQI-derived state information used in RL-based network and service management in dynamic vehicular environments. SCAR combines K-means clustering and radial basis function networks (RBFNs) to generate compact state representations from high-dimensional CQI measurements for subsequent RL-based scheduling decisions. To improve compression quality, simulated annealing with stochastic tunneling (SAST) is used to optimize both clustering and RBFN training, achieving approximately 10\% lower representation distortion than baseline approaches. 
When integrated with RL-based management, the compressed state representations increase the proportion of decision intervals operating within the NGMN-defined feasible region by more than 15 percentage points for the best-performing RL configuration compared with its uncompressed counterpart, while maintaining competitive computational runtime. The trade-off analysis further demonstrates that SCAR preserves high throughput while regulating fairness within the NGMN requirements, avoiding both unfair and excessively fair operating conditions.

By decoupling the RL state representation from raw CQI dimensionality, SCAR provides a compression-aware framework that is not inherently tied to a specific radio access technology. Although the framework was evaluated using a detailed system-level simulation environment, the present work focuses on state-space compression, its integration with RL-based management, and controlled performance analysis. Future work will investigate validation using real vehicular network traces and experimental environments, as well as integration with advanced learning architectures, including graph-based and attention-based models.

\vspace{-0.1in}
\section*{Acknowledgment}
This research was conducted as part of the UKIERI-SPARC project: "DigIT—Digital Twins for Integrated Transportation Platform", with grant number UKIERI-SPARC/01/23.


\ifCLASSOPTIONcaptionsoff
  \newpage
\fi
\vspace{-0.1in}

\bibliographystyle{IEEEtran}
\renewcommand\refname{References}
\bibliography{ReferenceList}

\end{document}